\documentclass[letterpaper]{article} 
\usepackage{aaai2027}  
\usepackage[hyphens]{url}  
\usepackage{graphicx} 
\usepackage{natbib}  
\usepackage{caption} 
\usepackage{algorithm}
\usepackage{algorithmic}

\usepackage{newfloat}
\usepackage{listings}
\DeclareCaptionStyle{ruled}{labelfont=normalfont,labelsep=colon,strut=off} 
\floatstyle{ruled}
\newfloat{listing}{tb}{lst}{}
\floatname{listing}{Listing}

\usepackage{booktabs}

\def\pred{\mathit{pred}}
\def\error{\mathit{error}}
\def\cond{\mathit{cond}}
\def\accept{\mathit{accept}}
\def\assign{\mathit{assign}}
\def\Inc{\mathit{Inc}}
\def\Pred{\mathit{Pred}}
\usepackage{amsmath}
\usepackage{amsfonts}
\usepackage{multirow}
\usepackage{arydshln}
\title{Adversarially Robust Abductive Fusion of \\
Pre-trained Transformer-based Perception Models}
\author{
	Mario Leiva\textsuperscript{\rm 1},
    Yue Ma\textsuperscript{\rm 2},
    Qinru Qiu\textsuperscript{\rm 2},
    Gerardo Simari\textsuperscript{\rm 1}, and
	Paulo Shakarian\textsuperscript{\rm 2}
}
\affiliations{
	\textsuperscript{\rm 1}DCIC, Universidad Nacional del Sur (UNS) \& ICIC (UNS-CONICET), Bah\'ia Blanca, Argentina\\
	\textsuperscript{\rm 2}Syracuse University, Syracuse, NY USA\\
    mario.leiva@cs.uns.edu.ar, yma183@syr.edu, qiqiu@syr.edu, gis@cs.uns.edu.ar, pashakar@syr.edu
}

\begin{document}

\maketitle

\begin{abstract}
Deploying pre-trained perception models in novel environments degrades their accuracy under distributional shift, and assembling them alone does not recover it: combiners such as majority voting trade recall for precision and are brittle to coordinated failures. Prior metacognitive methods learn logical rules that flag a model's errors, but rely on hand-authored domain-knowledge cues (object-size priors, segmentation masks) that do not transfer to genuinely novel scenes. We show that this metacognitive layer can be learned \emph{without any domain knowledge} by exploiting vector-space geometry: per-model \emph{Label Vector Pools} (LVP), built from each model's own training embeddings, yield error-detection rules from the geometry of detections relative to training-determined prototypes, reaching parity with domain-knowledge rules to within $0.002$ every F1 on test set. Because the approach remains neurosymbolic, these geometric rules share a single logical framework and can still be complemented by domain knowledge when available. We frame the fusion of multiple imperfect ViT-based detectors as a \emph{consistency-based abduction} problem solved at test time by an exact Integer Program (IP) and a polynomial-time heuristic. On an aerial-imagery benchmark of 15 weather-shifted test sets and six ViT detectors, our domain-knowledge-free layer matches the strongest majority-vote variant on clean data (within $0.005$ F1) and, unlike every majority-vote baseline, retains its performance under a coordinated label-flipping attack: at a $90\%$ flip rate it averages $0.42$ F1 versus $0.35$ for MV-Plurality (a $22\%$ relative gain) and attains the highest F1 on \emph{every} test set once the flip rate exceeds $0.4$.
\end{abstract}


\section{Introduction}
\label{sec:introduction}
Pre-trained perception models are now the default for classification and detection in images and video~\cite{han2021pre,radford2021learning}, and in practice they are deployed on the distribution they were trained on. We focus on the \emph{deployment in novel environments} setting: the operating conditions differ from training and no labeled data from target distribution is available. Emergency response after a disaster, or an aid mission to a remote region for which no representative imagery exists, are concrete examples---the scene is \emph{novel} relative to anything the models have seen.

Recent work~\cite{leiva26} has combined ideas from abducitve learning~\cite{dai2019abl} and metacognitive error detection rules~\cite{kricheli2024error} for test-time ensembling of vision models in a novel environment.  However, this approach requires the establishment of candidate metacognitive cues that the learner uses to derive rules.  This means that the system designer must have some a-priori domain knowledge on potential causes of error, which may lead to bespoke systems that to not generalize.  Further, while that work showed significant improvement over the baselines, it did not examine the case where some of the perceptual models may be subject to adversarial perturbations.


Here we extend the work of \cite{leiva26} by removing its two key limitations. The working hypothesis is the same: deploying \emph{more than one} model and reasoning over their joint (in)consistency can recover the recall that single-model error filtering throws away. We extend that line in two ways that we argue are necessary for novel and adversarial deployment.

\smallskip
\noindent\textbf{Contribution 1: A domain-knowledge-free metacognitive layer.}
We replace hand-authored domain cues with a learned signal that uses nothing but each model's own embeddings. Borrowing the \emph{Label Vector Pool} idea from continual learning with CLIP~\cite{Ma_2025_CVPR}, we represent each (model, class) by a small pool of prototype vectors obtained from the model's training detections, and train a light error detector on the distances of a new detection to these pools. The resulting ``LVP error probability'' is the only condition our error-detection rules use---no masks, no size priors, no scene semantics. This makes the entire metacognitive stack transferable to any new model or scene.

\smallskip
\noindent\textbf{Contribution 2: Robustness to attacks on majority voting.}
Majority voting is the canonical way to combine models, yet it is fragile since an adversary who can coordinate a minority of models can flip the vote. We study a \emph{coordinated label-flip} attack designed precisely to defeat plurality voting, and show that our abductive formulation degrades far more gracefully than any majority-vote variant because it accepts predictions based on cross-model \emph{consistency} rather than raw counts. 

\begin{figure*}[t]
\centering
\includegraphics[width=\linewidth]{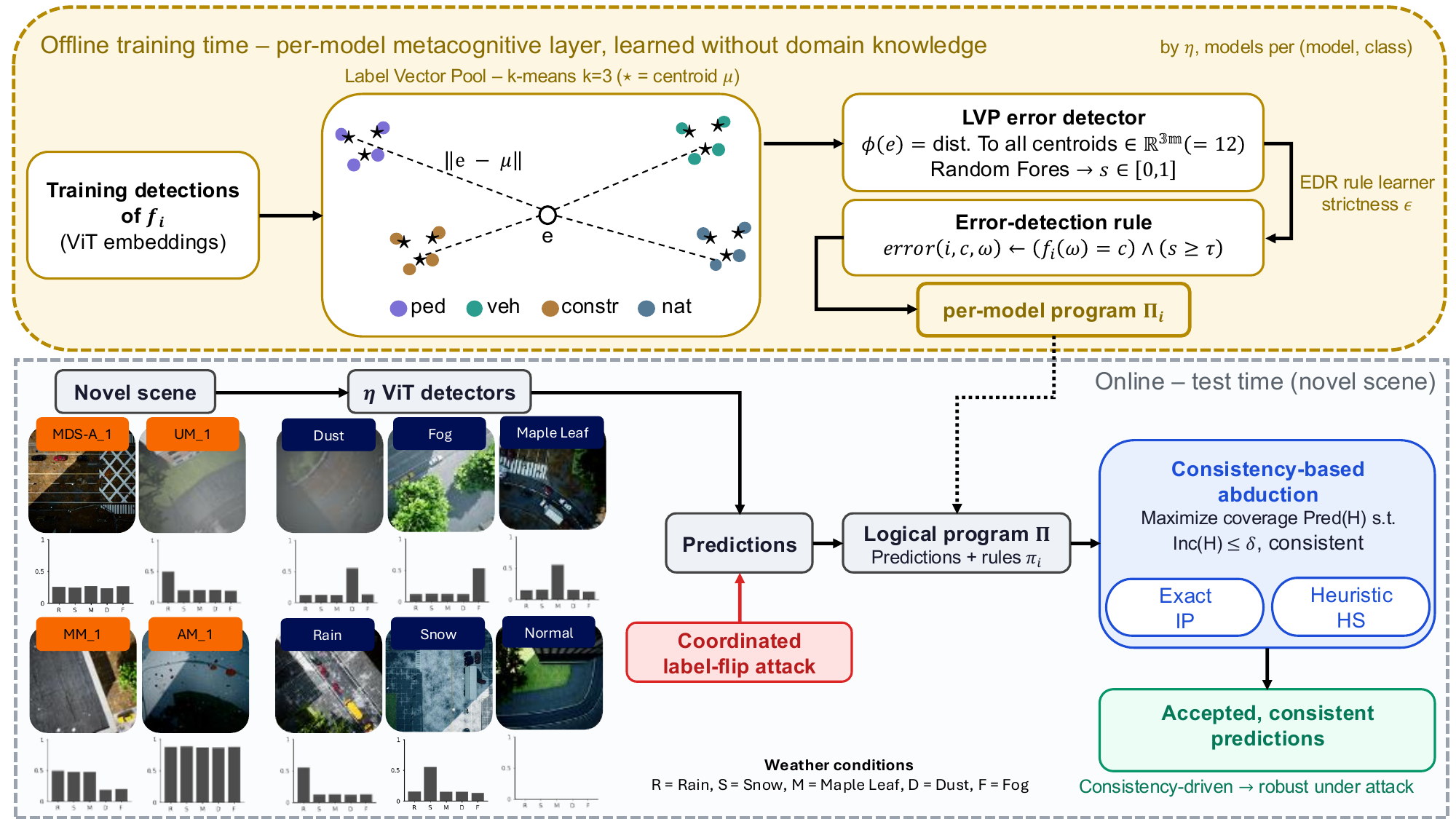}
\caption{\textbf{Overview.} \textit{(Top)} Offline, each of the $\eta$ pre-trained models builds a Label Vector Pool by clustering its training-detection embeddings with $k$-means into per-class prototype pools $\mathcal{P}_{i,c}=\{\mu_{i,c}^{1},\mu_{i,c}^{2},\mu_{i,c}^{3}\}$ ($\star$ = centroids). A new detection with embedding e is mapped to $\phi(e)\in\mathbb{R}^{3m}$ and a per-(model, class) Random Forest yields an error probability $s\in[0,1]$; thresholding $s\ge\tau$ via the EDR rule learner (strictness $\epsilon$) produces the error-detection program $\Pi_i$---using no domain knowledge and no target-distribution data. \textit{(Bottom)} At test time the $\eta$ models perceive a novel scene; their predictions, together with the per-model programs $\Pi_i$, form a logical program $\Pi$. Consistency-based abduction selects an accepted set maximizing coverage $\Pred(H)$ subject to $\Inc(H)\le\delta$, solved by an exact Integer Program (IP) or a polynomial-time Heuristic Search (HS), each with an optional tie-breaker (TB). Rules are learned independently per model from training data only, so there is no test-time leakage. Unlike majority voting, the combiner degrades gracefully under coordinated label-flip attacks because acceptance is consistency-driven, not count-driven.}
\label{fig:intro}
\end{figure*}

As summarized in Figure~\ref{fig:intro}, $\eta$ models perceive a novel scene; their predictions, together with LVP-derived metacognitive rules learned independently per model, are encoded in a logical program. We then abduce a subset of accepted predictions that maximizes coverage while keeping logical inconsistency below a budget. 
We derive an exact Integer Program (IP) and a scalable Heuristic Search (HS), each with an optional tie-breaker (TB). The rules are learned only from each model's training data, so there is no test-time leakage and no assumption about how models perform together.

\section{Related Work}
\label{sec:related_work}
\noindent\textbf{Error detection and correction rules.}  Prior work on error detection rules (EDR) relies on the assumption that there is prior knowledge on the potential causes of error - an assumption relaxed in this paper~\cite{xi2024rulebasederrordetectioncorrection,kricheli2024error}.  A recent use of EDR~\cite{leiva26} leverages this in an abductive framework for test-time ensembling, which we use as a baseline in this paper.  We differ in two ways: we operate over \emph{multiple} models, and our rule \emph{conditions} are derived from a learned LVP error detector rather than domain-specific predicates. We use only error \emph{detection} rules here, though correction rules fit the same framework.

\smallskip
\noindent\textbf{Abductive learning.}
ABL~\cite{dai2019abl} also uses abduction with domain knowledge, but to improve a model at training time and under the assumption that the test environment is not fully novel; the follow-up on new concepts~\cite{ijcai2021p250} extends the label scheme similarly to EDCR. We instead use abduction \emph{only at test time} and change the data distribution rather than the concept scheme. Classic consistency-based diagnosis~\cite{reiter198757,poole89} and the complexity of logic-based abduction~\cite{eiter95} inspire our formulation; to the best of our knowledge this machinery has not been applied to test-time perception over model ensembles.

\smallskip
\noindent\textbf{Label Vector Pool and prototypes.}
LVP-CLIP~\cite{Ma_2025_CVPR} keeps, per class/task, a \emph{pool} of label vectors rather than a single text embedding, enabling continual learning without forgetting. This is conceptually related to prototypical networks~\cite{snell2017prototypical}. We repurpose the pool idea not for classification but for \emph{error detection}: distances to per-class prototype pools are the features of a metacognitive classifier.

\smallskip
\noindent\textbf{Test-time training and adaptation.}
TTT~\cite{sun2020test} adapts the network itself at test time via self-supervision. This is complementary: a TTT-adapted model is just another pre-trained model in our framework, and several TTT variants could be combined through our abduction layer.

\smallskip
\noindent\textbf{Adversarial robustness of ensembles.}
Adversarial examples and poisoning are well studied~\cite{goodfellow2014explaining,madry2017towards,biggio2012poisoning,biggio2018wild,ming2024boosting}, and ensembles are not automatically robust~\cite{tramer2017ensemble}. We study a threat aimed specifically at the \emph{combiner}: a coordinated minority of models is made to agree on a wrong label, defeating plurality voting. 
Our consistency-based acceptance method is shown to be more resistant to this attack.

\section{Consistency-based Abduction}
\label{sec:problem}
We summarize the framework we build on; though the formulation follows~\cite{leiva26}, we keep it self-contained.

\smallskip
\noindent\textbf{Preliminaries.}
We consider object identification over perception data $\Omega$ with $\eta$ models $\mathcal{F}=\{f_1,\dots,f_\eta\}$ predicting over $m$ classes $\mathcal{C}=\{c_1,\dots,c_m\}$. Under the unique-name assumption\footnote{Implementation details for the unique-name assumption are in the supplementary material (Sec.~A).}, each object $\omega\in\Omega$ detected by a model yields a fact $f_i(\omega)=c_j$; the set of such facts (``observations'') is $O$. 
Because models disagree and make mistakes, we introduce $\accept(i,c)$, which is true when we choose to trust model $f_i$'s outputs of class $c$. The set of all acceptance atoms is $\mathcal{H}$; a subset $H\subseteq\mathcal{H}$ is a \emph{hypothesis}.

Each model $f_i$ carries a logic program $\Pi_i$ of metacognitive rules of the form:
\[
\error(i,c,\omega) \leftarrow (f_i(\omega)=c)\wedge \cond(\omega).
\]
I.e., if cue $\cond$ fires for $\omega$ and $f_i$ labeled it $c$, an error is suspected. 
The novelty here is that $\cond$ is the single LVP-derived predicate $\mathrm{LVP\_Error\_Probability\_GE}_\tau$, learned without domain knowledge. Helper program $\Pi_\textit{helper}$ contains
\[
\small
\assign(c,\omega)\leftarrow \neg\error(i,c,\omega)\wedge(f_i(\omega)=c)\wedge\accept(i,c)
\]
and a domain program $\Pi_\textit{dom}$ contains integrity constraints $\neg\assign(c',\omega)\leftarrow\assign(c,\omega)$ forbidding conflicting labels on one object. This integrity constraint is the \emph{only} domain knowledge used to judge consistency, and it is inherent to the task (the mutual exclusivity of class labels) rather than environment-specific expertise: it needs no segmentation masks, size priors, or other scene artifacts, and holds unchanged in any novel domain. All discriminative, scene-specific knowledge instead lives in the learned error-detection rules $\Pi_i$; keeping the consistency layer minimal is deliberate, and is what lets the same abduction machinery transfer across domains. 
We write $\Pi=\Pi_\textit{dom}\cup\Pi_\textit{helper}\cup(\bigcup_i\Pi_i)$; stratification and limited negation, in addition to the instance size, make inference tractable and monotonic, realized in PyReason~\cite{aditya2023pyreason}.

\smallskip
\noindent\textbf{Abduction problem.}
Following consistency-based abduction~\cite{eiter95,reggia91}, we seek $H\subseteq\mathcal{H}$ such that $H\cup O\cup\Pi$ is consistent. We allow a small, controlled amount of inconsistency: $\Inc(H)$ is the normalized number of ground rules in $\Pi_\textit{dom}$ not entailed by $(H\cup O\cup\Pi)\setminus\Pi_\textit{dom}$, and $\delta\in[0,1]$ bounds it. Among consistent hypotheses we prefer the most informative one: $\Pred(H)$ counts the $\assign(c,\omega)$ atoms entailed by the minimal model of $(H\cup O\cup\Pi)\setminus\Pi_\textit{dom}$. We \emph{maximize} $\Pred(H)$ because 
(i)~more assignments means fewer suspected errors on well-trained models, (ii)~we want high recall, and 
(iii)~$\Pi_\textit{helper}$ already guards against over-assignment. The problem is
\[
\max_{H\subseteq\mathcal{H}} \Pred(H)\,\text{s.t.}\, \Inc(H)\le\delta,\;
(H\cup O\cup\Pi)\setminus\Pi_\textit{dom}\text{ consist}.
\]
The rule-learner exposes a hyperparameter $\epsilon$ interpretable as the expected recall reduction from discarding flagged predictions; we sweep it and also set it heuristically.

\smallskip
\noindent\textbf{Integer Program (IP).}
With binary $A_{c,\omega}$ (object $\omega$ assigned class $c$), $\textit{Con}_{\omega,(c,c')}$ (conflict indicator), $\textit{Elim}_{f,c}$ (exclude model $f$'s class-$c$ output; $\textit{Elim}_{f,c}=0\Leftrightarrow
\accept(f,c)\in H$), $X_{\omega,f,c}$ (consider $(\omega,f,c)$), and constant $\pred_{f,c,\omega}$, we solve
\[
\max \sum_{\omega}\sum_{c}A_{c,\omega}
\]
subject to $X_{\omega,f,c}\le 1-\textit{Elim}_{f,c}$; $X_{\omega,f,c}\cdot\pred_{f,c,\omega}\le A_{c,\omega}$; $A_{c,\omega}\le\sum_f X_{\omega,f,c}\cdot\pred_{f,c,\omega}$;
$A_{c,\omega}+A_{c',\omega}-1\le\textit{Con}_{\omega,(c,c')}$ for $(c,c')\in IC$; $\sum_c A_{c,\omega}\ge 1$; and the global budget $\sum_{\omega}\sum_{(c,c')}\textit{Con}_{\omega,(c,c')}\le\delta\cdot\sum_{i,\omega}A_{i,\omega}$. The model has $O(N\cdot|\mathcal{F}|\cdot|\mathcal{C}|)$ variables and constraints ($N=|\Omega|$). It is NP-hard in general, but the locality of the conflict and consideration constraints makes the instances at our scale solvable in seconds.

\smallskip
\noindent\textbf{Heuristic Search (HS).}
Algorithm~\ref{alg:hs} greedly decides $\accept(f,c)$ per (model,class) pair: for each pair it picks the $\epsilon$ whose filtered predcition set most increases coverage while keeping global inconsistency $\le\delta$. HS runs in $O(|\mathcal{F}|\cdot|\mathcal{C}|\cdot|E_\textit{set}|)$, polynomial in the input.

\begin{algorithm}[t]
\caption{\small Heuristic Search (HS)}
\label{alg:hs}
\begin{algorithmic}[1]
\small
\STATE \textbf{Input:} $P_\textit{raw}$ (raw prediction tuples); $\delta$ (budget);
$E_\textit{set}$ ($\epsilon$ thresholds)
\STATE \textbf{Output:} $S_\textit{final}$ (accepted predictions)
\STATE $S_\textit{final}\leftarrow\emptyset$
\FOR{each model $f\in\mathcal{F}$, class $c\in\mathcal{C}$}
    \STATE $P_\textit{best}\leftarrow\emptyset$;\; $n_\textit{max}\leftarrow|S_\textit{final}|$
    \FOR{each $\epsilon\in E_\textit{set}$}
        \STATE $P_\textit{new}\leftarrow\textit{GetFilteredPreds}(f,c,\epsilon,P_\textit{raw})$
        \STATE $S_\textit{cand}\leftarrow S_\textit{final}\cup P_\textit{new}$
        \IF{$\textit{CalcIncon}(S_\textit{cand})\le\delta$ \AND $|S_\textit{cand}|>n_\textit{max}$}
            \STATE $P_\textit{best}\leftarrow P_\textit{new}$;\; $n_\textit{max}\leftarrow|S_\textit{cand}|$
        \ENDIF
    \ENDFOR
    \IF{$P_\textit{best}\neq\emptyset$} \STATE $S_\textit{final}\leftarrow S_\textit{final}\cup P_\textit{best}$ \ENDIF
\ENDFOR
\STATE \textbf{return} $S_\textit{final}$
\end{algorithmic}
\end{algorithm}

\smallskip
\noindent\textbf{Tie-Breaker (TB).}
Abduction can leave several admissible labels for an object. TB makes the output deterministic: for an object $\omega$ with multiple admisible labels it keeps the pair $(\omega,c)$ from the model with the highest confidence, yielding the IP+TB and HS+TB variants.

\section{Learning the Metacognitive Layer without Domain Knowledge}
\label{sec:lvp}
Here, we describe how each model's error-detection program $\Pi_i$ is learned from a \emph{Label Vector Pool} (LVP) instead of hand-authored domain predicates.

\smallskip
\noindent\textbf{Why no domain knowledge.}
The domain-knowledge (DK) conditions used by prior work are predicates such as ``the pedestrian box center lies on the street segmentation mask'' or ``the box area is far from the average class area in training.'' These require artifacts that a \emph{novel} environment does not provide (a segmentation map of the new scene, reliable size priors) and they are model-agnostic, so they cannot capture \emph{which model} is unreliable \emph{where}. We want a condition that (i)~needs only the model's own training data and (ii)~is specific to each (model, class).

\smallskip
\noindent\textbf{Label Vector Pool per (model, class).}
Each detection produced by a ViT-backbone detector comes with a $d$-dimensional embedding ($d=1024$). For a model $f_i$ and class~$c$, we take the embeddings of $f_i$'s \emph{training} detections of class~$c$ and cluster them with $k$-means ($k=3$), obtaining a pool $\mathcal{P}_{i,c}=\{\mu_{i,c}^{1},\mu_{i,c}^{2},
\mu_{i,c}^{3}\}$ of prototype vectors. The collection $\{\mathcal{P}_{i,c}\}_{c\in\mathcal{C}}$ is model $f_i$'s Label Vector Pool: a compact, multi-modal summary of how each class looks \emph{to that model under its own training condition}. Sub-clustering ($k>1$) matters because a class is rarely unimodal in embedding space (e.g., vehicles seen from different angles).

\smallskip
\noindent\textbf{LVP error detector.}
For a new detection with embedding $e$, we form the feature vector of Euclidean distances to every prototype of every class,
\[
\phi(e)=\big[\,\lVert e-\mu_{i,c}^{r}\rVert_2\,\big]_{c\in\mathcal{C},\,r\in\{1,2,3\}}
\in\mathbb{R}^{3m},
\]
i.e.\ $3\times 4=12$ features in our setting. A per-(model, class) Random Forest maps $\phi(e)$ to the probability that the detection is an \emph{error},
\[
s_{i,c}(e)=\Pr[\text{error}\mid \phi(e)]\in[0,1],
\]
where a training detection is labeled an error if it does not match a ground-truth object of the same class at $\mathrm{IoU}\ge 0.5$. Intuitively, a detection far from its predicted class's pool (or close to another class's pool) is suspicious. 
Score $s_{i,c}$ is what we call the \emph{LVP error probability}; it is the only component of our metacognitive conditions and uses no domain semantics.

\smallskip
\noindent\textbf{From scores to rules.}
We threshold the score into a family of monotone conditions $\cond_\tau(\omega)\equiv \big(s_{i,c}(e_\omega)\ge\tau\big)$ for a grid of $\tau$, and feed them to the EDCR rule learner~\cite{xi2024rulebasederrordetectioncorrection,kricheli2024error}. For each (model, class) and each strictness level $\epsilon$, the learner selects the condition(s) that yield error-detection rules meeting its precision criterion; $\epsilon$ is interpretable as the expected recall reduction. The output is the program $\Pi_i$ introduced in the Consistency-based Abduction section, where the suspected-error predicate is fired when the learned LVP threshold is exceeded. Because pools and detectors are built independently per model on each model's own training data, there is no test-time leakage and no cross-model coupling assumed a priori.
Empirically, $\epsilon$ acts as a monotone control: as it increases, the learner selects a lower effective threshold on the LVP error probability, flagging more detections and trading retained recall for higher per-model precision. This $\epsilon$ to threshold mapping differs across detectors, each reflecting its own error-score distribution, so a single $\epsilon$ induces model-specific operating points without any domain-specific tuning (a per-model curve is given in the supplementary material, Sec.~C).

\smallskip
\noindent\textbf{Putting it together.}
The full test-time pipeline is: (1)~each ViT detector emits boxes, labels, and embeddings (also a confidence score) on the novel scene; (2)~the LVP detector assigns each detection an error probability; (3)~the learner rules fire suspected errors; (4)~IP or HS abduces an accepted set under the inconsistency budget $\delta$; (5)~TB resolves residual ties. Steps 1--3 are the contribution of this section; steps 4--5 are the abduction machinery introduced earlier under Consistency-based Abduction.

\section{Experimental Setup}
\label{sec:experiments}

\smallskip
\noindent\textbf{Dataset.}
We use the Multiple Distribution Shift -- Aerial (MDS-A) dataset~\cite{ngu2025multipledistributionshift}, generated with AirSim~\cite{airsim2017fsr}, in the 15-test-set extension introduced in~\cite{leiva26}. Images are captured in a city environment under varying weather, with bounding boxes in four classes: \emph{pedestrian}, \emph{vehicle}, \emph{construction}, \emph{nature}. Six training conditions (\emph{rain, snow, fog, maple leaves, dust}, and a no-weather \emph{normal}) and a test suite of 15 sets spanning increasingly complex mixed-weather shifts are used: ids encode how many conditions share an intensity level: UM (unimodal), BM (bimodal), MM (most modes), AM (all modes), HUM (high unimodal). Figure~\ref{fig:intro} shows the conditions of the training sets for the~6 detectors and~4 of the~15 test sets (Figure~1 in the supp.\ material illustrates all conditions).

\smallskip
\noindent\textbf{Detection models.}
The six detectors use a plain Vision Transformer (ViT) backbone~\cite{dosovitskiy2020image} in the ViTDet configuration~\cite{li2022exploring}, initialized from MAE self-supervised pretraining~\cite{he2022masked} and paired with a Faster R-CNN~\cite{ren2015faster} detection head; all are fine-tuned with Detectron2~\cite{wu2019detectron2}. Each model is specialized to a single weather condition and trained in isolation, so that the mixed-weather test sets induce a genuine distributional shift. The embedding we use is a feature from each detector's \emph{own} detection head, and each model is fine-tuned end-to-end per weather condition, backbone and head together, with no frozen shared backbone and no adapters. Each model therefore has its own feature space, so the prototype pools $\mathcal{P}_{i,c}$ are model-specific and not directly comparable across models, which is why the metacognitive layer is built per model. Since each model specializes in one condition, the index $i$ denotes both the model $f_i$ and the weather domain it is trained for. All six share the same four output classes, and for every detection we take the $1024$-dimensional feature produced by the detection head's penultimate fully-connected layer as the embedding consumed by the LVP error detectors. Full training hyperparameters per-model are reported in the supplementary material (Sec.~B).

\smallskip
\noindent\textbf{Multimodal test set.}
Beyond MDS-A, we evaluate on VisDrone-DroneVehicle~\cite{sun2020drone}, a multimodal aerial dataset of paired RGB and infrared (IR) images (see Fig.~\ref{fig:drone:gt} for a sample pair). This probes a different use of the framework: the ensemble is two modality-specialized detectors (one RGB, one IR) rather than six weather specialized ones. We evaluate on a mixed test set of 250 RGB and 250 IR scenes drawn from disjoint location, so each detector sees both in- and out-of-domain inputs; with van and freight\_car too rare to learn, the pipeline runs on three effective classes (car, truck, bus). Preprocessing and sampling details are in the supplement (Sec.~H).

\begin{figure}[t]
  \centering
  \begin{tabular}{cc}
    \includegraphics[width=0.48\linewidth]{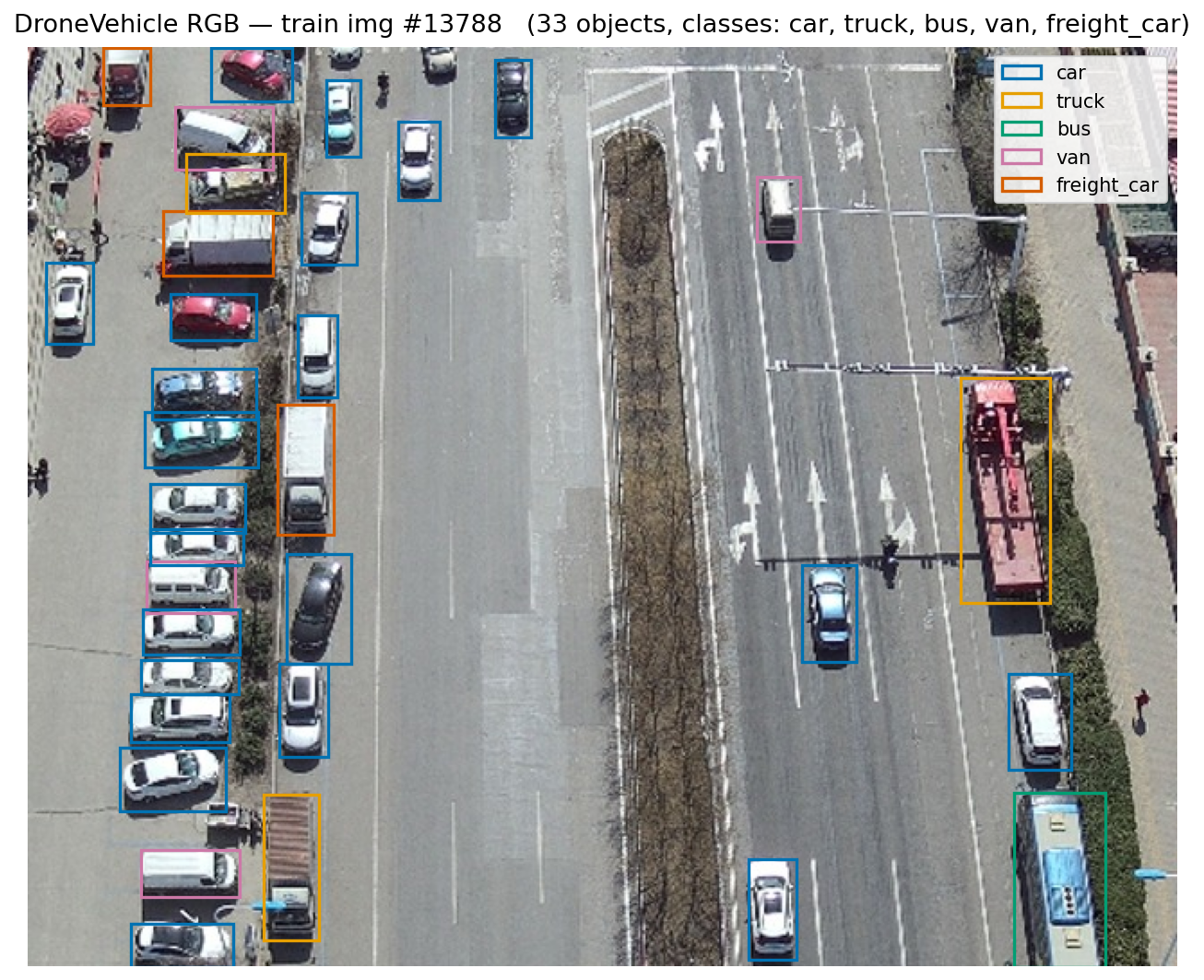} & \hspace*{-8pt}
    \includegraphics[width=0.47\linewidth]{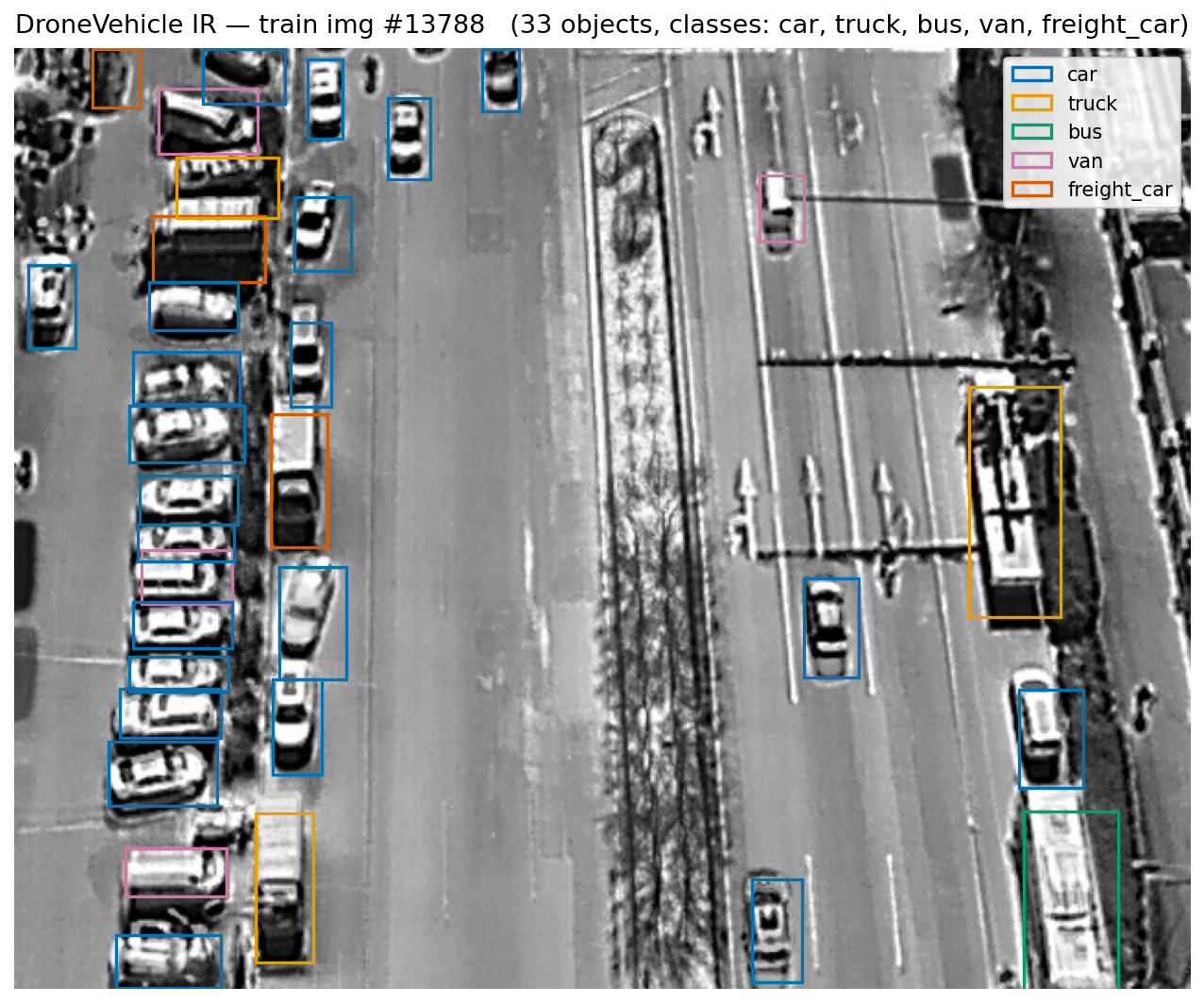} \\
  \end{tabular}
  \caption{DroneVehicle sample scene from the same physical location in
    RGB (left) and IR (right), with ground-truth bounding boxes
    color-coded by class.}
  \label{fig:drone:gt}
\end{figure}

\smallskip
\noindent\textbf{Baselines.}
We compare against per-model references--- \emph{MacroAVG} (mean over models) and \emph{Best Model} (best single model)---and against all majority-vote (MV) variants: \emph{MV-Plurality} (plurality among active detectors), \emph{MV-Plurality-NoTB} (plurality without the confidence tie-break), \emph{MV-Quorum} (requires $\lfloor N/2\rfloor+1$ active detectors), and \emph{MV-Strict} (plurality across all models, treating ``no detection'' as a vote). Our methods are IP, IP+TB, HS and HS+TB. Also, as a first step after the model detections, ghost detections (detections that did not match any object in the ground truth) were eliminated, in order to work only with the true FPs: detections with incorrect labels.

\smallskip
\noindent\textbf{Hardware and software.}
The detectors were fine-tuned on a single NVIDIA T4 GPU (Google Colab); every other stage---detection inference, LVP error detection, EDR rule firing, and the IP/HS abduction---runs on CPU only. We use a single-socket server with an AMD EPYC 9755  running Ubuntu 24.04 and Python 3.12, with no GPU. Test scenarios are processed in parallel across cores via multiprocessing. Logical deduction is implemented in PyReason~\cite{aditya2023pyreason}, and the integer program is solved with the CBC solver through the PuLP modeling library.
\section{Results}
\label{sec:results}

\smallskip
\noindent\textbf{Clean-data performance.}
Table~\ref{tab:main} reports F1 and Accuracy on a representative subset of the test sets under clean (unattacked) conditions and the average over all 15 test sets. On these scenarios, our IP+TB and HS+TB are statistically \emph{on par} with the strongest ensemble baseline, MV-Plurality, while outperforming the per-model references and the precision-biased MV variants. However, we note that our method reaches parity with the best baselines but does not rely on voting, which is susceptible to correlation-based attacks (as described later).

\begin{table}[t]
\small
\centering
\setlength{\tabcolsep}{3pt}
\caption{F1 / Accuracy on a representative subset of test sets for best individual model, MV-Plurality (MV-P) and our approach using the abductive framework of~\cite{leiva26}. The Average row is over all 15 datasets. Accuracy is the detection Jaccard index, $\mathrm{Acc}=F1/(2-F1)$. The full 15-scenario table appears in the supplement (Sec.~D).}
\label{tab:main}
\begin{tabular}{l|cc|cc|cc|cc}
\toprule
\multirow{2}{*}{\textbf{Test Set}}
 & \multicolumn{2}{c|}{\textbf{Best}}
 & \multicolumn{2}{c|}{\textbf{MV-P}}
 & \multicolumn{2}{c|}{\textbf{IP+TB}}
 & \multicolumn{2}{c}{\textbf{HS+TB}} \\
 & F1 & Acc & F1 & Acc & F1 & Acc & F1 & Acc \\
\midrule
MDS-A\_1 & 0.70 & 0.54 & \textbf{0.76} & \textbf{0.61} & 0.75 & 0.60 & 0.75 & 0.60 \\
UM\_1    & 0.64 & 0.47 & \textbf{0.72} & \textbf{0.56} & \textbf{0.72} & \textbf{0.56} & \textbf{0.72} & \textbf{0.56} \\
MM\_1    & 0.64 & 0.47 & \textbf{0.73} & \textbf{0.58} & \textbf{0.73} & \textbf{0.58} & \textbf{0.73} & \textbf{0.58} \\
AM\_1    & 0.30 & 0.17 & \textbf{0.34} & \textbf{0.21} & \textbf{0.34} & \textbf{0.21} & \textbf{0.34} & \textbf{0.21} \\
\hdashline
Avg. (15 tests)  & 0.55 & 0.38 & \textbf{0.64} & \textbf{0.47} & 0.63 & 0.46 & \textbf{0.64} & \textbf{0.47} \\
\midrule
DroneVehicle & \textbf{0.68} & \textbf{0.51} & 0.67 & 0.50 & 0.67 & 0.50 & 0.67 & 0.50 \\
\bottomrule
\end{tabular}
\end{table}

\begin{figure}[t]
    \centering
    \includegraphics[width=\linewidth]{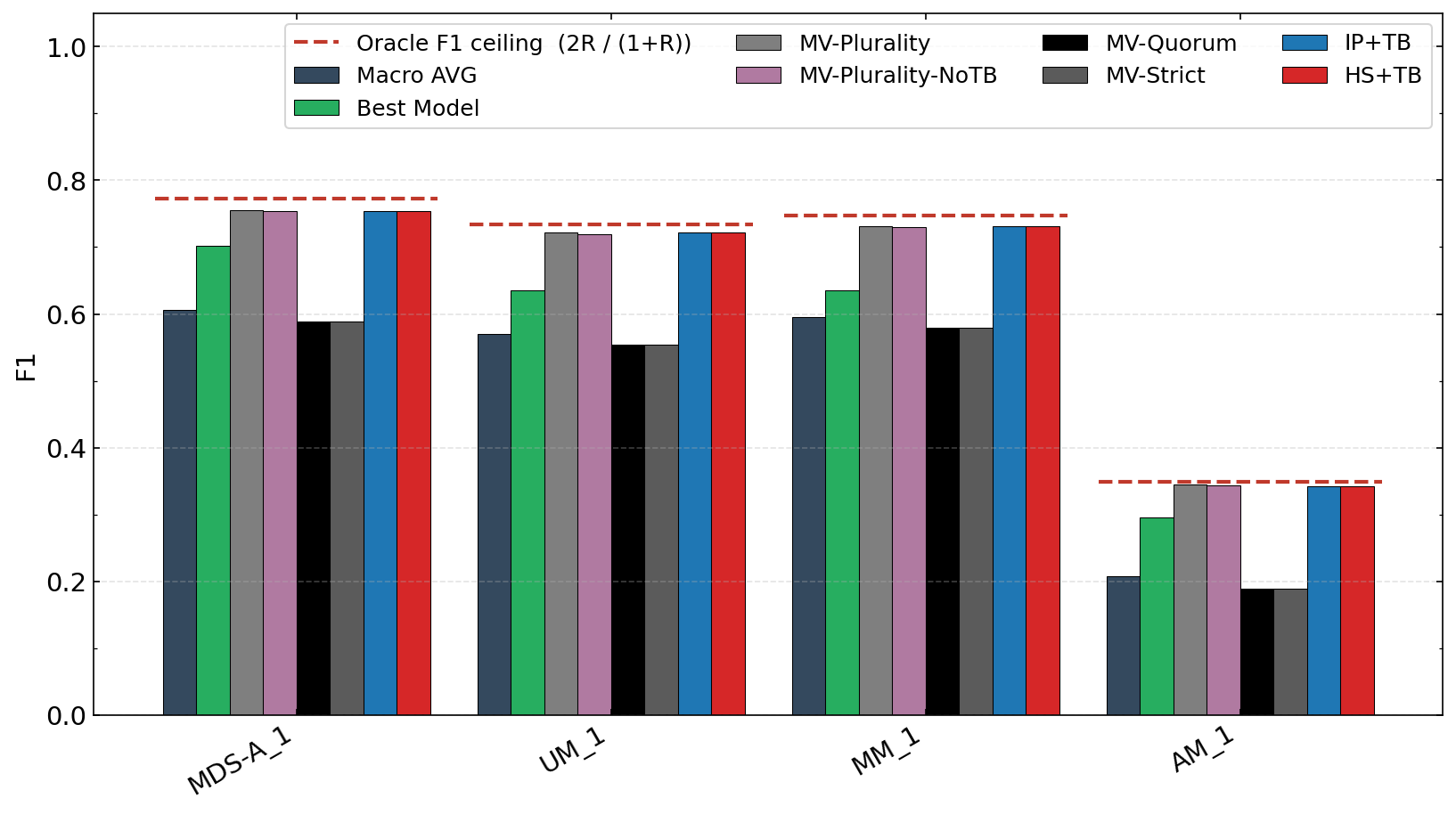}
    \caption{F1 of each method per test set with the oracle-recall ceiling drawn
    above each group. Methods that select among existing predictions cannot exceed
    the dashed line; IP+TB sits closest to it.}
    \label{fig:ceiling}
\end{figure}

\smallskip
\noindent\textbf{Ceiling analysis.}
\label{sec:results:ceiling}
For each test set we compute two ceilings over methods that only \emph{select} among existing predictions: the \emph{detection coverage} (fraction of ground-truth objects seen by at least one model) and the \emph{oracle recall} (fraction seen with the \emph{correct} class by at least one model), with corresponding oracle F1. Figure~\ref{fig:ceiling} plots each method against these ceilings on the same representative subset used in Table~\ref{tab:main} (AM\_1, BM\_1, MDS-A\_1, MM\_1, UM\_1); the remaining 10 test sets are variations of the same five families (AM\_2/3, BM\_2/3, MDS-A\_2/3, MM\_2/3, UM\_2/3, HUM\_1) and behave qualitatively the same. The gap between the precision-based MV variants (MV-Quorum, MV-Strict) and the oracle is large, whereas MV-Plurality and our methods approach the oracle.

\begin{figure}[t]
    \centering
    \includegraphics[width=\linewidth]{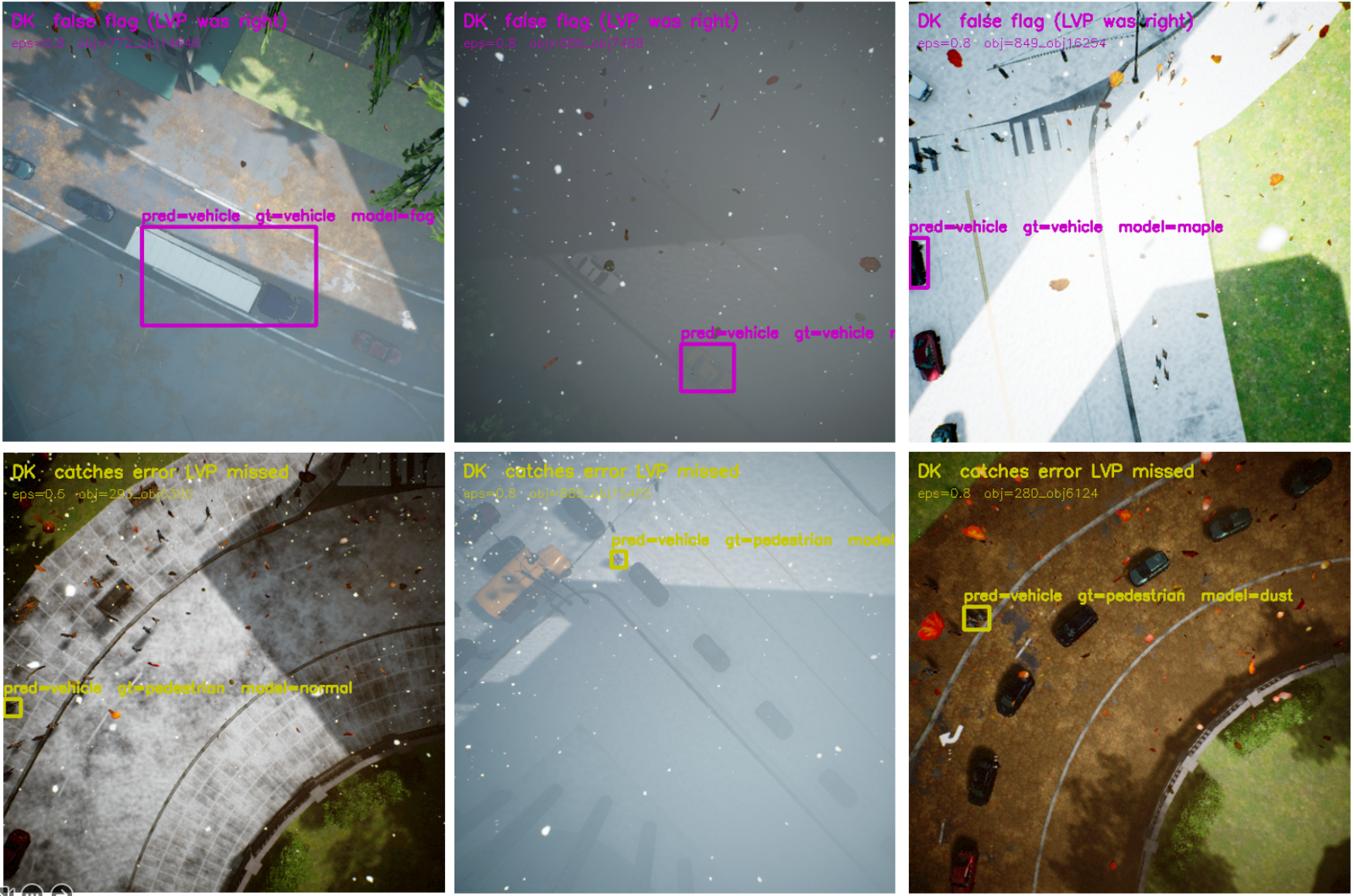}
    \caption{Detections where the LVP and DK rules disagree. \emph{Top:} errors the DK rules miss but the LVP rules catch, typically model-specific embedding-space anomalies that no hand-authored cue captures; \emph{Bottom:} the converse, errors the LVP rules miss but the DK rules catch. Although LVP needs no domain knowledge and matches DK on downstream F1, the per-detection view shows the two flag different failures.}
    \label{fig:lvp_dk_examples}
\end{figure}

\smallskip
\noindent\textbf{LVP vs.\ domain-knowledge rules.}
\label{sec:results:lvpdk}
We compare the LVP-based rules against the domain-knowledge (DK) rules used by prior work, holding the abduction stage fixed. The headline results are in Table~\ref{tab:lvp_vs_dk}: \emph{the two reach essentially the same downstream F1}---the largest absolute difference across the five test sets is $|\Delta|=0.0018$ (i.e., below $0.002$ F1)---yet the LVP variant requires \emph{no} scene artifacts: no segmentation masks, no class-specific size priors, only each model's own training embeddings. This method reaches parity with a domain-knowledge baseline \emph{without} using any domain knowledge---the metacognitive layer can be made portable to genuinely novel scenes at no measurable cost in downstream accuracy. The per-detection disagreement analysis in Figure~\ref{fig:lvp_dk_examples} adds a finer view: although the aggregate F1 is identical, the two signals catch different failures---LVP flags model-specific embedding-space anomalies that no hand authored cue captures (top), while DK catches geometric/contextual mistakes that the embedding alone does not (bottom). The two are complementary, and combining them is a natural direction for future work.

\begin{table}[t]
\centering
\small
\caption{Downstream F1 with LVP vs.\ DK conditions (abduction fixed). LVP requires no domain artifacts and reaches the same downstream F1 as DK (larges gap $<0.002$).}
\label{tab:lvp_vs_dk}
\begin{tabular}{l|cc|c}
\toprule
\textbf{Test Set} & \textbf{IP+TB (DK)} & \textbf{IP+TB (LVP)} & \textbf{$\Delta$ ($\times 10^{-4}$)} \\
\midrule
MDS-A\_1 & 0.7534 & 0.7538 & +4 \\ 
UM\_1 & 0.7206 & 0.7214 & +8 \\ 
MM\_1 & 0.7314 & 0.7307 & -7 \\ 
AM\_1 & 0.3438 & 0.3420 & -18 \\ 
\bottomrule
\end{tabular}
\end{table}

\smallskip
\noindent\textbf{Robustness to coordinated attacks.}
We assume an attacker who can perturb predictions to defeat the combiner but cannot retrain detectors. With probability $p$ per ground-truth object, the attacker samples three models and one wrong class and forces those three models to output that class for the object. The flips are \emph{coordinated}: three models agree on the same wrong label, impacting voting-based techniques that have a latent independence assumption among the models.

\smallskip
\noindent\textbf{Multimodal test.}
On clean data the multimodal ensemble is the hardest case for abduction: RGB and IR were trained on paired scenes and rarely disagree (inter-detector inconsistency $\approx 0.0002$), so IP+TB and HS+TB reduce to the majority-vote solution and tie MV-Plurality at $F_1=0.672$, just below the best single detector (IR, $F_1=0.68$). Under the coordinated label-flip attack, this regime becomes the best case: IP+TB and HS+TB stay above every baseline for all $p\ge 0.1$, reaching $F_1=0.395$ and $0.410$ at $p=0.9$ against $0.376$ for MV-Plurality, while MV-Plurality-NoTB collapses. They also stay closest to the clean-data ceiling across the whole sweep (Figure~\ref{fig:adv}, right); the full $F_1$-vs-$p$ curves and per-method numbers are in the supplementary material (Sec.~H). The disagreements the abduction layer resolves arise naturally from six weather detectors on MDS-A, but only under attack on the two correlated modality detectors; the same mechanism yields clean-data wins on one dataset and adversarial wins on the other.


\smallskip
\noindent\textbf{Results.}
Table~\ref{tab:adv_f1} lists F1 across all 15 test sets at five attack rates that span the non-trivial regime ($p\in\{0.2,0.4,0.6,0.8,0.9\}$), for the strongest representative of each method family---\emph{Best} (best per-model baseline, ahead of Macro AVG), \emph{MV-Plurality} (best majority-vote variant, ahead of MV-Plurality-NoTB, MV-Quorum and MV-Strict), and our \emph{IP+TB} and \emph{HS+TB}. Clean-data values ($p=0$) are already in Table~\ref{tab:main}; the full grid with all ten methods and all $p\in\{0.0,\dots,0.9\}$ is in the supplementary material (Sec.~E). 
The pattern is unambiguous: at $p=0.2$ IP+TB and MV-Plurality are still close (often tied), but from $p=0.4$ onward IP+TB takes the lead on \emph{every} test set, and by $p=0.9$ MV-Plurality has lost roughly $0.10$--$0.15$ F1 while IP+TB has barely moved. HS+TB follows the same trend and stays comfortably above the MV baseline. The take-away is that the gap to the count-based combiner grows monotonically with $p$, which is what we should expect if acceptance is consistency-driven rather than vote-driven.

\begin{table*}[t]
\centering
\scriptsize
\setlength{\tabcolsep}{3pt}
\caption{F1 across all 15 test sets under the coordinated label-flip attack, at attack rates $p\in\{0.2,0.4,0.6,0.8,0.9\}$. Mean over 3 seeds. \textbf{Bold}: highest F1 within each (test set, $p$) cell (ties highlighted together). As soon as the attack becomes non-trivial, IP+TB takes the lead on \emph{every} test set, and the gap to MV-P widens monotonically with $p$---direct evidence that consistency-based abduction is far more attack-resistant than count-based combiners. Acc derivable as $\mathrm{Acc}=F1/(2-F1)$.}
\label{tab:adv_f1}
\resizebox{\textwidth}{!}{%
\begin{tabular}{l|cccc|cccc|cccc|cccc|cccc}
\toprule
\multirow{2}{*}{\textbf{Test Set}}
 & \multicolumn{4}{c|}{$p=0.2$}
 & \multicolumn{4}{c|}{$p=0.4$}
 & \multicolumn{4}{c|}{$p=0.6$}
 & \multicolumn{4}{c|}{$p=0.8$}
 & \multicolumn{4}{c}{$p=0.9$} \\
 & Best & MV-P & IP+TB & HS+TB
 & Best & MV-P & IP+TB & HS+TB
 & Best & MV-P & IP+TB & HS+TB
 & Best & MV-P & IP+TB & HS+TB
 & Best & MV-P & IP+TB & HS+TB \\
\midrule
MDS-A\_1 & 0.63 & 0.68 & \textbf{0.69} & 0.68 & 0.57 & 0.61 & \textbf{0.63} & \textbf{0.63} & 0.49 & 0.53 & 0.58 & \textbf{0.59} & 0.43 & 0.45 & \textbf{0.54} & 0.50 & 0.39 & 0.42 & \textbf{0.52} & 0.46 \\
MDS-A\_2 & 0.64 & 0.69 & 0.69 & \textbf{0.70} & 0.57 & 0.61 & 0.64 & \textbf{0.65} & 0.50 & 0.54 & 0.58 & \textbf{0.59} & 0.43 & 0.46 & \textbf{0.54} & 0.53 & 0.40 & 0.42 & \textbf{0.52} & 0.48 \\
MDS-A\_3 & 0.41 & 0.51 & \textbf{0.52} & 0.51 & 0.37 & 0.45 & \textbf{0.47} & 0.45 & 0.32 & 0.39 & \textbf{0.43} & 0.41 & 0.28 & 0.34 & \textbf{0.40} & 0.35 & 0.26 & 0.31 & \textbf{0.39} & 0.33 \\
UM\_1    & 0.57 & \textbf{0.65} & \textbf{0.65} & \textbf{0.65} & 0.51 & 0.58 & \textbf{0.59} & 0.58 & 0.45 & 0.51 & \textbf{0.54} & 0.52 & 0.39 & 0.43 & \textbf{0.50} & 0.44 & 0.35 & 0.40 & \textbf{0.48} & 0.41 \\
UM\_2    & 0.55 & 0.63 & 0.63 & \textbf{0.64} & 0.48 & 0.56 & \textbf{0.58} & 0.56 & 0.42 & 0.49 & \textbf{0.53} & 0.51 & 0.37 & 0.42 & \textbf{0.49} & 0.44 & 0.33 & 0.38 & \textbf{0.47} & 0.40 \\
UM\_3    & 0.49 & 0.58 & \textbf{0.59} & 0.58 & 0.43 & 0.51 & \textbf{0.53} & 0.52 & 0.38 & 0.45 & \textbf{0.49} & 0.46 & 0.33 & 0.39 & \textbf{0.45} & 0.41 & 0.30 & 0.36 & \textbf{0.43} & 0.38 \\
BM\_1    & 0.52 & 0.60 & \textbf{0.61} & 0.60 & 0.47 & 0.54 & \textbf{0.55} & 0.54 & 0.41 & 0.47 & \textbf{0.50} & 0.48 & 0.35 & 0.41 & \textbf{0.46} & 0.43 & 0.32 & 0.37 & \textbf{0.44} & 0.40 \\
BM\_2    & 0.44 & \textbf{0.52} & \textbf{0.52} & \textbf{0.52} & 0.39 & 0.46 & \textbf{0.47} & 0.46 & 0.34 & 0.40 & \textbf{0.42} & 0.41 & 0.30 & 0.35 & \textbf{0.39} & 0.36 & 0.27 & 0.31 & \textbf{0.37} & 0.33 \\
BM\_3    & 0.50 & 0.58 & \textbf{0.59} & 0.58 & 0.44 & 0.52 & \textbf{0.53} & \textbf{0.53} & 0.39 & 0.45 & \textbf{0.48} & 0.47 & 0.33 & 0.39 & \textbf{0.44} & 0.41 & 0.30 & 0.35 & \textbf{0.42} & 0.38 \\
MM\_1    & 0.57 & \textbf{0.66} & \textbf{0.66} & \textbf{0.66} & 0.51 & 0.58 & \textbf{0.60} & 0.59 & 0.44 & 0.50 & \textbf{0.55} & 0.52 & 0.38 & 0.44 & \textbf{0.52} & 0.46 & 0.35 & 0.40 & \textbf{0.50} & 0.44 \\
MM\_2    & 0.42 & \textbf{0.49} & \textbf{0.49} & \textbf{0.49} & 0.37 & 0.43 & \textbf{0.44} & 0.43 & 0.33 & 0.38 & \textbf{0.40} & 0.39 & 0.28 & 0.32 & \textbf{0.36} & 0.33 & 0.26 & 0.29 & \textbf{0.34} & 0.30 \\
MM\_3    & 0.59 & 0.66 & \textbf{0.67} & \textbf{0.67} & 0.53 & 0.59 & \textbf{0.62} & 0.60 & 0.46 & 0.51 & \textbf{0.56} & 0.53 & 0.40 & 0.44 & \textbf{0.52} & 0.47 & 0.36 & 0.40 & \textbf{0.50} & 0.43 \\
AM\_1    & 0.27 & \textbf{0.31} & 0.30 & \textbf{0.31} & 0.24 & \textbf{0.28} & \textbf{0.28} & \textbf{0.28} & 0.21 & 0.24 & \textbf{0.26} & 0.24 & 0.18 & 0.21 & \textbf{0.24} & 0.21 & 0.16 & 0.19 & \textbf{0.22} & 0.19 \\
AM\_2    & 0.35 & \textbf{0.42} & \textbf{0.42} & \textbf{0.42} & 0.31 & 0.37 & \textbf{0.38} & \textbf{0.38} & 0.27 & 0.32 & \textbf{0.35} & 0.33 & 0.24 & 0.28 & \textbf{0.33} & 0.29 & 0.22 & 0.26 & \textbf{0.31} & 0.27 \\
HUM\_1   & 0.51 & \textbf{0.59} & \textbf{0.59} & \textbf{0.59} & 0.46 & 0.52 & \textbf{0.55} & 0.53 & 0.40 & 0.46 & \textbf{0.51} & 0.47 & 0.34 & 0.39 & \textbf{0.46} & 0.40 & 0.32 & 0.36 & \textbf{0.45} & 0.37 \\
\bottomrule
\end{tabular}%
}
\end{table*}

The left panel of Figure~\ref{fig:adv} shows a complementary view. The figure aggregates over all 15 test sets and plots the \emph{gap to the oracle F1 ceiling} per method as $p$ grows: this is the headline robustness picture---IP+TB stays closest to the oracle across the entire attack range, MV-Plurality and MV-Plurality-NoTB lose ground sharply, and the precision-based MV variants pull away from the oracle even faster. Acceptance based on cross-model \emph{consistency} (our methods) is therefore far more attack-resistant than acceptance based on raw vote counts (MV). Our methods never fall below the baselines once the attack is non-trivial: for $p\ge0.4$, Table~\ref{tab:adv_f1} shows IP+TB best or tied on every one of the 15 test sets, so the overlapping bands in Figure~\ref{fig:adv} reflect how much headroom varies across scenes, rather than cases where a baseline overtakes our methods.


\begin{figure}[t]
    \centering
    \includegraphics[width=\linewidth]{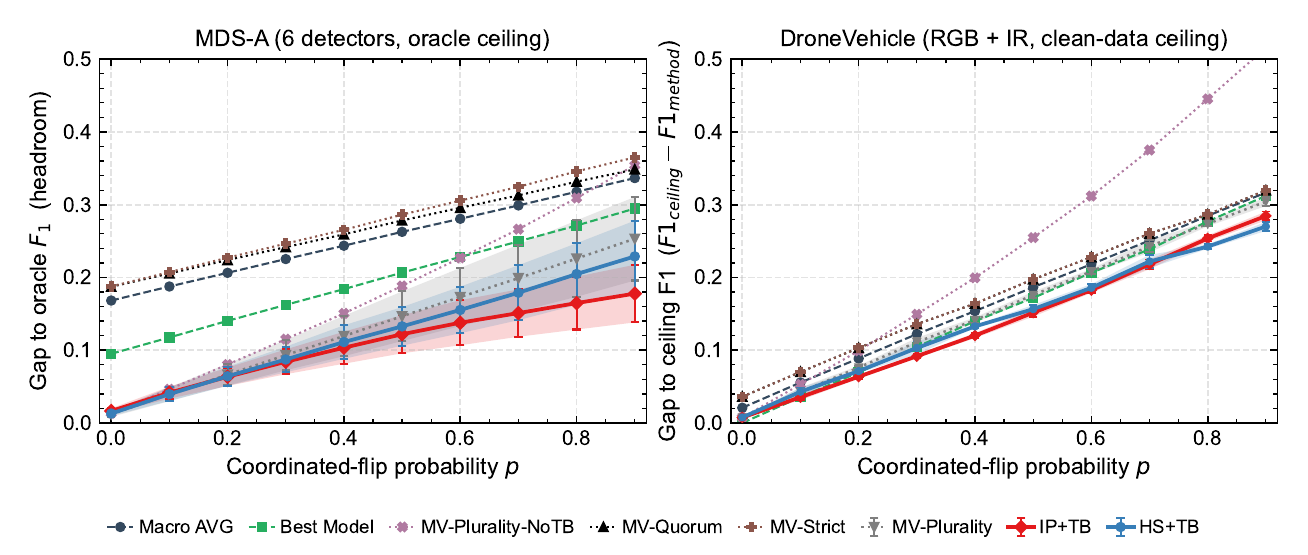}
    \caption{Adversarial coordinated label-flip: gap to the F1 ceiling per method as flip probability $p$ grows (lower is better); panel titles and $y$-axes name the ceiling used in each. \emph{(Left)} MDS-A over the 15 test sets, bands $\pm1$ std across test sets. \emph{(Right)} Multimodal test ($k=1$) over 3 adversary seeds, bands $\pm1$ std over seeds; its ceiling is a fixed constant, so the informative signal is the ordering among methods, not the absolute gap. IP+TB and HS+TB stay closest to the ceiling in both; full multimodal curves in the supp.\ material (Sec.~H).}
    \label{fig:adv}
\end{figure}

\begin{figure}[t]
    \centering
    \includegraphics[width=0.85\linewidth]{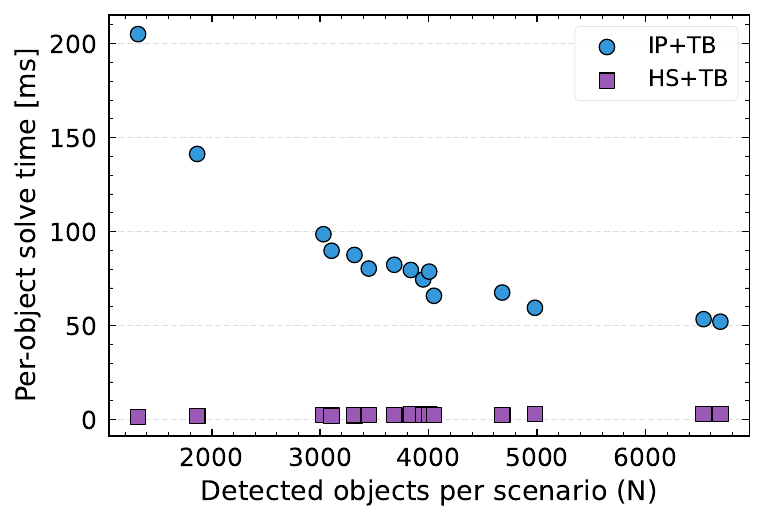}
    \caption{Wall-clock cost of IP+TB and HS+TB on the 15 clean test sets ($p=0$), one point scenario. Per-object solve time: HS+TB is essentially flat at $1.4$--$3.0$\,ms regardless of scenario size, matching the polynomial bound.}
    \label{fig:runtime}
\end{figure}

\smallskip
\noindent\textbf{Runtime.}
Figure~\ref{fig:runtime} presents the wall-clock cost of IP+TB and HS+TB on the 15 clean test sets ($p=0$). HS+TB completes a full scenario in $1.9$--$19.7$\,s, with per-object solve time pinned in a narrow $1.4$--$3.0$\,ms band across the entire range of scenario sizes (1.3k--6.7k detected objects), consistent with its $O(|\mathcal{F}|\cdot|\mathcal{C}|\cdot|E_\textit{set}|)$ analysis. IP+TB is roughly $30\times$ slower in absolute terms ($264$--$349$\,s per scenario) but its per-object cost actually \emph{decreases} from $0.21$\,s on the smallest scenario to $0.005$\,s on the largest, because the fixed CBC overhead amortizes over more objects: at our scale, the IP+TB approach is driven by setup, not by combinatorial growth. 
Both methods thus remain practical end-to-end---HS+TB for tight latency budgets, IP+TB when the (small) gain in F1 justifies the extra cost in terms of running time.

\section{Conclusions and Future Work}
\label{sec:conclusions-fw}

We showed that the metacognitive layer of a multi-model perception system can be learned \emph{without domain knowledge}, using only per-model Label Vector Pools, and that framing model integration as consistency-based abduction yields a combiner that is both competitive on clean data and robust to a coordinated attack that breaks majority voting. Future work includes richer deduction rules for alternative assignment sets, automatic FP-detection without an oracle, finer joint exploration of $\epsilon$ and $\delta$, and runtime optimization for real-time deployment. 
Furthermore, one promising direction for future work is combining the domain knowledge (DK-EDR) and LVP (LVP-EDR) versions of EDR, as they both work in a rule-based framework. In the supplementary material (Sec.~G), we report that the two approaches identify different sources of error, making them complementary. However, this complementary nature also has the effect of reducing true positives due to increases in error recall, and we found that combining the approaches did not meaningfully impact F1 (also reported in the supplementary material, Sec.~G).

\section{Acknowledgments}
Funded by ARO grant W911NF-24-1-0007 and experiments were performed with compute resources supported by ARO DURIP and equipment donated by AMD.

\bibliography{aaai2027}

@inproceedings{leiva26,
  author       = {Mario A. Leiva and
                  Noel Ngu and
                  Joshua Shay Kricheli and
                  Aditya Taparia and
                  Ransalu Senanayake and
                  Paulo Shakarian and
                  Nathaniel D. Bastian and
                  John Corcoran and
                  Gerardo I. Simari},
  editor       = {Sven Koenig and
                  Chad Jenkins and
                  Matthew E. Taylor},
  title        = {Consistency-based Abductive Reasoning over Perceptual Errors of Multiple
                  Pre-trained Models in Novel Environments},
  booktitle    = {Proceedings of {AAAI}},
  pages        = {19216--19223},
  publisher    = {{AAAI}~Press},
  year         = {2026},
  url          = {https://doi.org/10.1609/aaai.v40i23.38996},
  doi          = {10.1609/AAAI.V40I23.38996},
  bibsource    = {dblp computer science bibliography, https://dblp.org}
}

@InProceedings{Ma_2025_CVPR,
    author    = {Ma, Yue and Ren, Huantao and Wang, Boyu and Jin, Jingang and Velipasalar, Senem and Qiu, Qinru},
    title     = {LVP-CLIP: Revisiting CLIP for Continual Learning with Label Vector Pool},
    booktitle = {Proceedings of the IEEE/CVF Conference on Computer Vision and Pattern Recognition (CVPR) Workshops},
    month     = {June},
    year      = {2025},
    pages     = {231-240}
}

@book{reggia91, title={Abductive inference models for diagnostic problem-solving},
author ={Yun Peng and James A. Reggia}, 
publisher={Springer-Verlag},
year=1990
}

@article{eiter95,
author = {Eiter, Thomas and Gottlob, Georg},
title = {The complexity of logic-based abduction},
year = {1995},
issue_date = {Jan. 1995},
publisher = {Association for Computing Machinery},
address = {New York, NY, USA},
volume = {42},
number = {1},
issn = {0004-5411},
url = {https://doi.org/10.1145/200836.200838},
doi = {10.1145/200836.200838},
journal = {J. ACM},
month = jan,
pages = {3–42},
numpages = {40}
}

@misc{xi2024rulebasederrordetectioncorrection,
      title={Rule-Based Error Detection and Correction to Operationalize Movement Trajectory Classification}, 
      author={Bowen Xi and Kevin Scaria and Divyagna Bavikadi and Paulo Shakarian},
      year={2024},
      eprint={2308.14250},
      archivePrefix={arXiv},
      primaryClass={cs.LG},
      url={https://arxiv.org/abs/2308.14250}, 
}

@inproceedings{kricheli2024error,
  title={Error detection and constraint recovery in hierarchical multi-label classification without prior knowledge},
  author={Kricheli, Joshua Shay and Vo, Khoa and Datta, Aniruddha and Ozgur, Spencer and Shakarian, Paulo},
  booktitle={Proceedings {CIKM}},
  pages={3842--3846},
  year={2024}
}

@inproceedings{ngu2025multipledistributionshift,
      title={Multiple Distribution Shift -- Aerial (MDS-A): A Dataset for Test-Time Error Detection and Model Adaptation}, 
      author={Noel Ngu and Aditya Taparia and Gerardo I. Simari and Mario Leiva and Jack Corcoran and Ransalu Senanayake and Paulo Shakarian and Nathaniel D. Bastian},
    booktitle = {{AAAI} Spring Symposium},
    year={2025},
      eprint={2502.13289},
      archivePrefix={arXiv},
      primaryClass={cs.LG},
      url={https://arxiv.org/abs/2502.13289}, 
}

@inproceedings{airsim2017fsr,
  author = {Shital Shah and Debadeepta Dey and Chris Lovett and Ashish Kapoor},
  title = {AirSim: High-Fidelity Visual and Physical Simulation for Autonomous Vehicles},
  year = {2017},
  booktitle = {Field and Service Robotics},
  eprint = {arXiv:1705.05065},
  url = {https://arxiv.org/abs/1705.05065}
}

@article{dosovitskiy2020image,
  title={An image is worth 16x16 words: Transformers for image recognition at scale},
  author={Dosovitskiy, Alexey and Beyer, Lucas and Kolesnikov, Alexander and Weissenborn, Dirk and Zhai, Xiaohua and Unterthiner, Thomas and Dehghani, Mostafa and Minderer, Matthias and Heigold, Georg and Gelly, Sylvain and others},
  journal={arXiv preprint arXiv:2010.11929},
  year={2020}
}

@inproceedings{li2022exploring,
  title={Exploring plain vision transformer backbones for object detection},
  author={Li, Yanghao and Mao, Hanzi and Girshick, Ross and He, Kaiming},
  booktitle={Proceedings of {ECCV}},
  pages={280--296},
  year={2022},
  organization={Springer}
}

@inproceedings{he2022masked,
  title={Masked autoencoders are scalable vision learners},
  author={He, Kaiming and Chen, Xinlei and Xie, Saining and Li, Yanghao and Doll{\'a}r, Piotr and Girshick, Ross},
  booktitle={Proceedings of the IEEE/CVF conference on computer vision and pattern recognition},
  pages={16000--16009},
  year={2022}
}

@article{ren2015faster,
  title={Faster {R-CNN}: Towards real-time object detection with region proposal networks},
  author={Ren, Shaoqing and He, Kaiming and Girshick, Ross and Sun, Jian},
  journal={Advances in neural information processing systems},
  volume={28},
  year={2015}
}

@misc{wu2019detectron2,
  author =       {Yuxin Wu and Alexander Kirillov and Francisco Massa and
                  Wan-Yen Lo and Ross Girshick},
  title =        {Detectron2},
  howpublished = {\url{https://github.com/facebookresearch/detectron2}},
  year =         {2019}
}

@article{aditya2023pyreason,
  title={PyReason: Software for Open World Temporal Logic},
  author={Aditya, Dyuman and Mukherji, Kaustuv and Balasubramanian, Srikar and Chaudhary, Abhiraj and Shakarian, Paulo},
  journal={arXiv preprint arXiv:2302.13482},
  year={2023}
}

@article{han2021pre,
  title={Pre-trained models: Past, present and future},
  author={Han, Xu and Zhang, Zhengyan and Ding, Ning and Gu, Yuxian and Liu, Xiao and Huo, Yuqi and Qiu, Jiezhong and Yao, Yuan and Zhang, Ao and Zhang, Liang and others},
  journal={AI open},
  volume={2},
  pages={225--250},
  year={2021},
  publisher={Elsevier}
}

@inproceedings{radford2021learning,
  title={Learning transferable visual models from natural language supervision},
  author={Radford, Alec and Kim, Jong Wook and Hallacy, Chris and Ramesh, Aditya and Goh, Gabriel and Agarwal, Sandhini and Sastry, Girish and Askell, Amanda and Mishkin, Pamela and Clark, Jack and others},
  booktitle={International conference on machine learning},
  pages={8748--8763},
  year={2021},
  organization={PmLR}
}

@inproceedings{dai2019abl,
  title     = {Bridging Machine Learning and Logical Reasoning by Abductive Learning},
  author    = {Dai, Wang-Zhou and Xu, Qiuling and Yu, Yang and Zhou, Zhi-Hua},
  booktitle = {Advances in Neural Information Processing Systems (NeurIPS)},
  year      = {2019}
}

@inproceedings{ijcai2021p250,
  title     = {Abductive Learning with Ground Knowledge Base},
  author    = {Cai, Le-Wen and Dai, Wang-Zhou and Huang, Yu-Xuan and Li, Yu-Feng and Muggleton, Stephen and Jiang, Yuan},
  booktitle = {Proceedings of {IJCAI}},
  publisher = {International Joint Conferences on Artificial Intelligence Organization},
  editor    = {Zhi-Hua Zhou},
  pages     = {1815--1821},
  year      = {2021},
  month     = {8},
  note      = {Main Track},
  doi       = {10.24963/ijcai.2021/250},
  url       = {https://doi.org/10.24963/ijcai.2021/250},
}

@inproceedings{poole89,
author = {Poole, David},
title = {Normality and faults in logic-based diagnosis},
year = {1989},
publisher = {Morgan Kaufmann Publishers Inc.},
address = {San Francisco, CA, USA},
booktitle = {Proceedings of the 11th International Joint Conference on Artificial Intelligence - Volume 2},
pages = {1304–1310},
numpages = {7},
location = {Detroit, Michigan},
series = {IJCAI'89}
}

@article{reiter198757,
title = {A theory of diagnosis from first principles},
journal = {Artificial Intelligence},
volume = {32},
number = {1},
pages = {57-95},
year = {1987},
issn = {0004-3702},
doi = {https://doi.org/10.1016/0004-3702(87)90062-2},
url = {https://www.sciencedirect.com/science/article/pii/0004370287900622},
author = {Raymond Reiter}
}

@article{snell2017prototypical,
  title={Prototypical networks for few-shot learning},
  author={Snell, Jake and Swersky, Kevin and Zemel, Richard},
  journal={Advances in neural information processing systems},
  volume={30},
  year={2017}
}

@inproceedings{sun2020test,
  title={Test-time training with self-supervision for generalization under distribution shifts},
  author={Sun, Yu and Wang, Xiaolong and Liu, Zhuang and Miller, John and Efros, Alexei and Hardt, Moritz},
  booktitle={International conference on machine learning},
  pages={9229--9248},
  year={2020},
  organization={PMLR}
}

@article{goodfellow2014explaining,
  title={Explaining and harnessing adversarial examples},
  author={Goodfellow, Ian J and Shlens, Jonathon and Szegedy, Christian},
  journal={arXiv preprint arXiv:1412.6572},
  year={2014}
}

@article{madry2017towards,
  title={Towards deep learning models resistant to adversarial attacks},
  author={Madry, Aleksander and Makelov, Aleksandar and Schmidt, Ludwig and Tsipras, Dimitris and Vladu, Adrian},
  journal={arXiv preprint arXiv:1706.06083},
  year={2017}
}

@article{biggio2012poisoning,
  title={Poisoning attacks against support vector machines},
  author={Biggio, Battista and Nelson, Blaine and Laskov, Pavel},
  journal={arXiv preprint arXiv:1206.6389},
  year={2012}
}

@inproceedings{biggio2018wild,
  title={Wild patterns: Ten years after the rise of adversarial machine learning},
  author={Biggio, Battista and Roli, Fabio},
  booktitle={Proceedings of the 2018 ACM SIGSAC conference on computer and communications security},
  pages={2154--2156},
  year={2018}
}

@inproceedings{
ming2024boosting,
title={Boosting the Transferability of Adversarial Attack on Vision Transformer with Adaptive Token Tuning},
author={Di Ming and Peng Ren and Yunlong Wang and Xin Feng},
booktitle={The Thirty-eighth Annual Conference on Neural Information Processing Systems},
year={2024},
url={https://openreview.net/forum?id=sNz7tptCH6}
}

@article{tramer2017ensemble,
title={Ensemble adversarial training: Attacks and defenses},
author={Tram{\`e}r, Florian and Kurakin, Alexey and Papernot, Nicolas and Goodfellow, Ian and Boneh, Dan and McDaniel, Patrick},
journal={arXiv preprint arXiv:1705.07204},
year={2017}
}

@article{sun2020drone,
title={Drone-based RGB-Infrared Cross-Modality Vehicle Detection via Uncertainty-Aware Learning},
author={Sun, Yiming and Cao, Bing and Zhu, Pengfei and Hu, Qinghua},
journal={IEEE Transactions on Circuits and Systems for Video Technology},
year={2022},
volume={},
number={},
pages={1-1},
doi={10.1109/TCSVT.2022.3168279}
}

\newpage
\section{Supplementary Material}
\subsection{Dataset: The 15-Test-Set MDS-A Benchmark}
\begin{figure*}[t]
    \centering
    \includegraphics[width=0.98\linewidth]{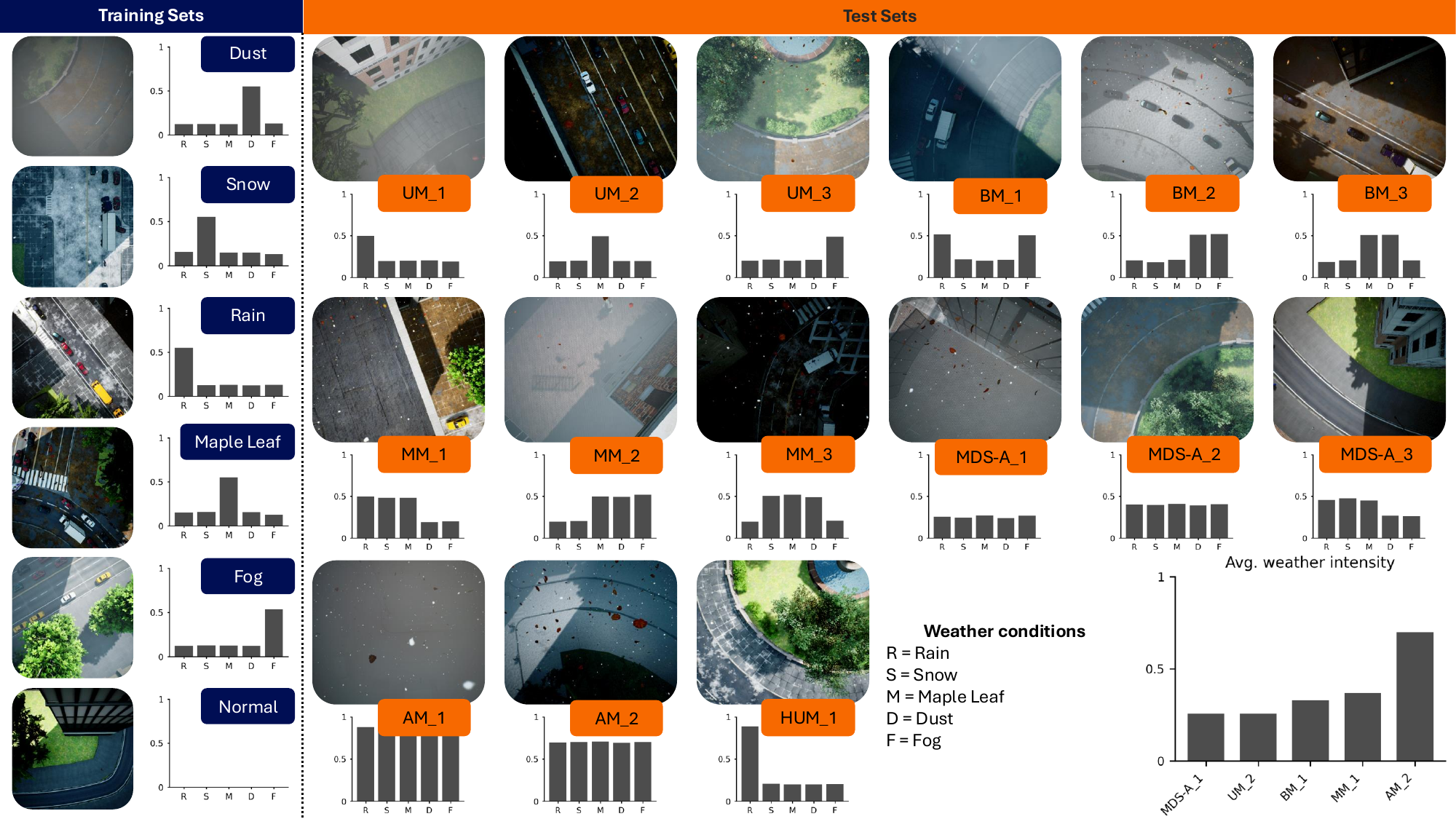}
    \caption{Example AirSim views under different weather conditions and the
    weather-intensity distribution of the corresponding test sets.}
    \label{fig:dataset}
\end{figure*}

We use the Multiple Distribution Shift -- Aerial (MDS-A) dataset~\cite{ngu2025multipledistributionshift}, generated with AirSim~\cite{airsim2017fsr}, in the 15-test-set extension introduced by Leiva et al.~\cite{leiva26}. Images are captured in a city environment under varying weather, with bounding boxes in four classes: \emph{pedestrian}, \emph{vehicle}, \emph{construction}, \emph{nature}. Six training conditions (\emph{rain,snow,fog,maple leaves, dust}, and a no-weather \emph{normal}) and a test suite of 15 sets spanning increasingly complex mixed-weather shifts are used. Figure~\ref{fig:dataset} illustrates the conditions.

\subsection{A. Unique-Name Assumption: Implementation}
\label{app:una}
We instantiate the unique-name assumption through the ensemble-matching step: a
detection is associated with a ground-truth object when their boxes overlap at
$\mathrm{IoU}\ge 0.5$; among multiple candidates from the same model we keep the
highest-IoU box \emph{regardless of predicted class}, so that genuine inter-model
disagreements are preserved. Detections that match no ground-truth object are kept
as their own object identifiers (FP-detections).

\subsection{B. Detection Models: Training Configuration}
\label{app:detectors}
The six detectors follow the ViTDet recipe~\cite{li2022exploring}: a plain ViT-Base backbone pretrained with MAE~\cite{he2022masked} and a Faster R-CNN detection head, fine-tuned per weather condition with Detectron2. Table~\ref{tab:detcfg} lists the shared training configuration.

\begin{table}[h]
\centering
\small
\caption{Detector training configuration.}
\label{tab:detcfg}
\begin{tabular}{l|c}
\toprule
\textbf{Setting} & \textbf{Value} \\
\midrule
Backbone               & ViT-Base (plain) \\
Detection head         & Faster R-CNN (ViTDet) \\
Pretraining            & MAE (ImageNet-1K) \\
Optimizer / base LR    & AdamW / $10^{-4}$ \\
Batch size / iters     & 1 / 3000 \\
Input resolution       & $1024\times1024$ (LSJ) \\
Test confidence thr.   & 0.7 \\
Embedding dim.         & 1024 \\
\bottomrule
\end{tabular}
\end{table}

\subsection{C. LVP Error Detector: Training Details}
\label{app:lvp}
Per (model, class) Random Forest are trained on the distance features $\phi(e)$ ($3$ sub-centroids $\times\,4$ classes $=12$ dims), with hyperparameters selected by 3-fold cross-validation. Table~\ref{tab:clf} reports the error-detector Precision, Recall and F1 averaged across the four classes and across the 15 test scenarios, one row per backbone-model condition.

\begin{table}[h]
\centering
\small
\caption{LVP error-detector metrics per model, averaged across all classes and all test scenarios (mean $\pm$ std).}
\label{tab:clf}
\begin{tabular}{l|ccc}
\toprule
\textbf{Backbone model} & \textbf{Precision} & \textbf{Recall} & \textbf{F1} \\
\midrule
dust    & .69 $\pm$ .09 & .17 $\pm$ .03 & .28 $\pm$ .04 \\
fog     & .72 $\pm$ .07 & .24 $\pm$ .02 & .36 $\pm$ .03 \\
maple   & .85 $\pm$ .08 & .25 $\pm$ .06 & .39 $\pm$ .07 \\
rain    & .74 $\pm$ .04 & .15 $\pm$ .02 & .25 $\pm$ .03 \\
snow    & .61 $\pm$ .05 & .13 $\pm$ .02 & .21 $\pm$ .02 \\
normal  & .88 $\pm$ .04 & .25 $\pm$ .02 & .39 $\pm$ .03 \\
\midrule
\textbf{Average} & .74 $\pm$ 0.06 & 0.19 $\pm$ 0.03 & 0.31 $\pm$ 0.04 \\
\bottomrule
\end{tabular}
\end{table}

\begin{figure}[t]
    \centering
    \includegraphics[width=\linewidth]{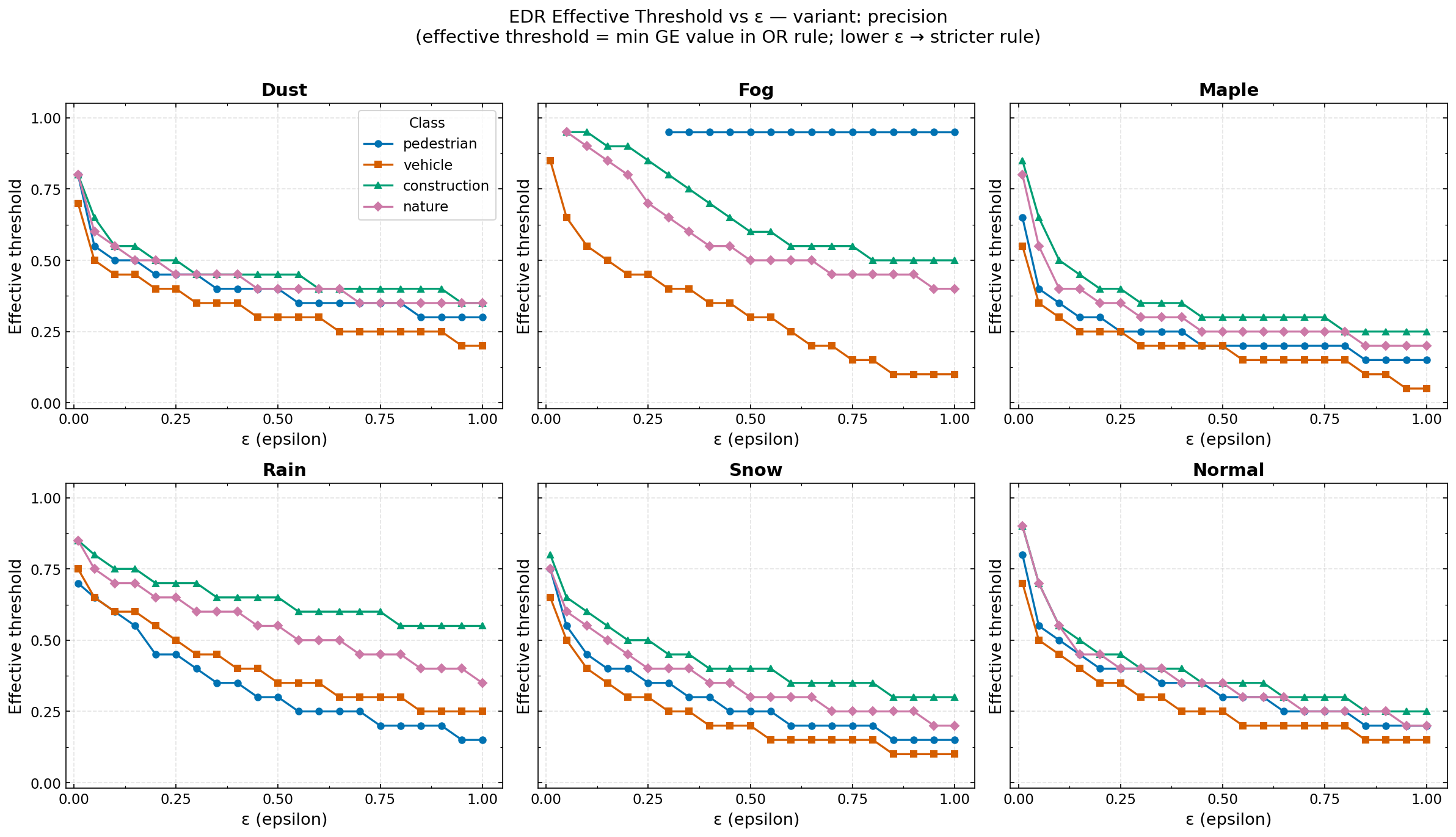}
    \caption{Effective LVP error-probability threshold as a function of the rule-learner strictness $\epsilon$, one curve per detector. Larger $\epsilon$ lowers the effective threshold, yielding more error detections and lower retained recall. The per-model curves differ in slope and range, showing that the same $\epsilon$ produces different operating points for different detectors.}
    \label{fig:eff_thr_vs_eps}
\end{figure}

\noindent\textbf{Hyperparameter Study: Rule-Learner Strictness ($\epsilon$)}
We also examine how the strictness hyperparameter $\epsilon$ of the rule learner shapes the LVP-derived programs. Recall that $\epsilon$ is interpretable as the expected recall reduction from discarding flagged predictions. Figure~\ref{fig:eff_thr_vs_eps} shows the \emph{effective threshold} on the LVP error probability picked by the learner as a function of $\epsilon$, separately per model: as $\epsilon$ grows, the effective threshold decreases monotonically. A lower threshold means the rule fires on lower error scores, which in turn flags more detections as errors---more aggressive filtering, more downstream recall reduction, and (on the per-model side) higher precision. The slope and operating range differ across the six models, reflecting how each detector's error-score distribution responds to its own LVP pool; this is what makes the rule-learning stage \emph{model-specific without being domain-specific}.

\subsection{D. Full Per-Scenario Results}
\label{app:fulltables}
Table~\ref{app:tab:full} gives F1 and Accuracy for every method on all 15 test sets, the full version of Table 1 in the main paper.

\begin{table*}[t]
\centering
\scriptsize
\setlength{\tabcolsep}{4pt}
\caption{F1 / Accuracy across the 15 test sets (clean data). In
bold: the strongest baseline (MV-Plurality in every set) and our two methods
(IP+TB, HS+TB), i.e.\ the top-performing cluster---our methods land within
$\approx0.005$ F1 of the best baseline on every set. Columns: Avg.\ (macro average
over models), Best (best single model), MV-P (MV-Plurality), MV-P$^\dagger$
(MV-Plurality without tie-break), MV-Q (MV-Quorum), MV-S (MV-Strict), IP+TB, HS+TB.
Accuracy is the detection Jaccard index, $\mathrm{Acc}=F1/(2-F1)$.}
\label{app:tab:full}
\resizebox{\textwidth}{!}{%
\begin{tabular}{l|cc|cc|cc|cc|cc|cc|cc|cc}
\toprule
\multirow{2}{*}{\textbf{Test Set}}
 & \multicolumn{2}{c|}{\textbf{Avg.}}
 & \multicolumn{2}{c|}{\textbf{Best}}
 & \multicolumn{2}{c|}{\textbf{MV-P}}
 & \multicolumn{2}{c|}{\textbf{MV-P$^\dagger$}}
 & \multicolumn{2}{c|}{\textbf{MV-Q}}
 & \multicolumn{2}{c|}{\textbf{MV-S}}
 & \multicolumn{2}{c|}{\textbf{IP+TB}}
 & \multicolumn{2}{c}{\textbf{HS+TB}} \\
 & F1 & Acc & F1 & Acc & F1 & Acc & F1 & Acc & F1 & Acc & F1 & Acc & F1 & Acc & F1 & Acc \\
\midrule
MDS-A\_1 & 0.61 & 0.43 & 0.70 & 0.54 & \textbf{0.76} & \textbf{0.61} & 0.75 & 0.61 & 0.59 & 0.42 & 0.59 & 0.42 & \textbf{0.75} & \textbf{0.60} & \textbf{0.75} & \textbf{0.60} \\
MDS-A\_2 & 0.61 & 0.44 & 0.72 & 0.56 & \textbf{0.77} & \textbf{0.62} & 0.77 & 0.62 & 0.59 & 0.42 & 0.59 & 0.42 & \textbf{0.76} & \textbf{0.61} & \textbf{0.77} & \textbf{0.62} \\
MDS-A\_3 & 0.41 & 0.25 & 0.46 & 0.30 & \textbf{0.56} & \textbf{0.39} & 0.56 & 0.39 & 0.38 & 0.24 & 0.38 & 0.24 & \textbf{0.57} & \textbf{0.40} & \textbf{0.56} & \textbf{0.39} \\
UM\_1    & 0.57 & 0.40 & 0.64 & 0.47 & \textbf{0.72} & \textbf{0.56} & 0.72 & 0.56 & 0.55 & 0.38 & 0.55 & 0.38 & \textbf{0.72} & \textbf{0.56} & \textbf{0.72} & \textbf{0.56} \\
UM\_2    & 0.56 & 0.39 & 0.61 & 0.43 & \textbf{0.70} & \textbf{0.54} & 0.70 & 0.54 & 0.54 & 0.37 & 0.54 & 0.37 & \textbf{0.70} & \textbf{0.54} & \textbf{0.70} & \textbf{0.54} \\
UM\_3    & 0.47 & 0.31 & 0.54 & 0.37 & \textbf{0.65} & \textbf{0.48} & 0.65 & 0.48 & 0.46 & 0.30 & 0.46 & 0.30 & \textbf{0.65} & \textbf{0.48} & \textbf{0.65} & \textbf{0.48} \\
BM\_1    & 0.50 & 0.33 & 0.58 & 0.41 & \textbf{0.67} & \textbf{0.51} & 0.67 & 0.51 & 0.48 & 0.32 & 0.48 & 0.32 & \textbf{0.67} & \textbf{0.50} & \textbf{0.67} & \textbf{0.51} \\
BM\_2    & 0.40 & 0.25 & 0.49 & 0.32 & \textbf{0.58} & \textbf{0.41} & 0.58 & 0.41 & 0.38 & 0.24 & 0.38 & 0.24 & \textbf{0.58} & \textbf{0.41} & \textbf{0.58} & \textbf{0.41} \\
BM\_3    & 0.50 & 0.33 & 0.55 & 0.38 & \textbf{0.65} & \textbf{0.48} & 0.65 & 0.48 & 0.48 & 0.32 & 0.48 & 0.32 & \textbf{0.65} & \textbf{0.48} & \textbf{0.65} & \textbf{0.48} \\
MM\_1    & 0.59 & 0.42 & 0.64 & 0.47 & \textbf{0.73} & \textbf{0.58} & 0.73 & 0.57 & 0.58 & 0.41 & 0.58 & 0.41 & \textbf{0.73} & \textbf{0.58} & \textbf{0.73} & \textbf{0.58} \\
MM\_2    & 0.37 & 0.23 & 0.47 & 0.31 & \textbf{0.54} & \textbf{0.37} & 0.54 & 0.37 & 0.35 & 0.21 & 0.35 & 0.21 & \textbf{0.54} & \textbf{0.37} & \textbf{0.54} & \textbf{0.37} \\
MM\_3    & 0.59 & 0.42 & 0.65 & 0.49 & \textbf{0.74} & \textbf{0.58} & 0.74 & 0.58 & 0.57 & 0.40 & 0.57 & 0.40 & \textbf{0.73} & \textbf{0.58} & \textbf{0.74} & \textbf{0.58} \\
AM\_1    & 0.21 & 0.12 & 0.30 & 0.17 & \textbf{0.34} & \textbf{0.21} & 0.34 & 0.21 & 0.19 & 0.10 & 0.19 & 0.10 & \textbf{0.34} & \textbf{0.21} & \textbf{0.34} & \textbf{0.21} \\
AM\_2    & 0.31 & 0.18 & 0.39 & 0.24 & \textbf{0.47} & \textbf{0.31} & 0.47 & 0.30 & 0.28 & 0.17 & 0.28 & 0.16 & \textbf{0.46} & \textbf{0.30} & \textbf{0.47} & \textbf{0.30} \\
HUM\_1   & 0.50 & 0.33 & 0.57 & 0.40 & \textbf{0.66} & \textbf{0.49} & 0.65 & 0.49 & 0.48 & 0.31 & 0.48 & 0.31 & \textbf{0.65} & \textbf{0.48} & \textbf{0.65} & \textbf{0.49} \\
\bottomrule
\end{tabular}%
}
\end{table*}

\subsection{E. Aggregated Adversarial Results (all methods, all $p$)}
\label{app:add_agg}
Table~\ref{tab:adv_agg} reports F1 averaged across the 15 test sets, for each of the 10 methods at all 10 attack rates $p\in\{0.0,\dots,0.9\}$. This complements Table~3 in the main paper, which shows per-scenario values for the four headline methods at five attack rates: here we trade per-scenario detail for full coverage of the method space (the four MV variants plus the no-TB versions of IP and HS that we omitted from the body) and of the attack-rate grid. Means are computed by first averaging across the 3 seeds per scenario, then across the 15 scenarios; the reported standard deviation is the across-scenario variability. \textbf{Bold}: top mean F1 per column. 

\begin{table*}[t]
\centering
\scriptsize
\setlength{\tabcolsep}{3pt}
\caption{F1 under the coordinated label-flip attack, averaged over the 15 test sets (mean $\pm$ across-scenario std; per-scenario values are themselves the mean over 3 seeds). \textbf{Bold}: top mean F1 per column. MV-P-NoTB abbreviates MV-Plurality-NoTB.}
\label{tab:adv_agg}
\resizebox{\textwidth}{!}{%
\begin{tabular}{l|cccccccccc}
\toprule
\textbf{Method} & $p=0.0$ & $p=0.1$ & $p=0.2$ & $p=0.3$ & $p=0.4$ & $p=0.5$ & $p=0.6$ & $p=0.7$ & $p=0.8$ & $p=0.9$ \\
\midrule
Macro AVG          & 0.48\,$\pm$.12 & 0.46\,$\pm$.11 & 0.43\,$\pm$.11 & 0.41\,$\pm$.10 & 0.38\,$\pm$.10 & 0.36\,$\pm$.09 & 0.34\,$\pm$.08 & 0.31\,$\pm$.08 & 0.29\,$\pm$.07 & 0.26\,$\pm$.07 \\
Best Model         & 0.55\,$\pm$.12 & 0.53\,$\pm$.11 & 0.50\,$\pm$.10 & 0.47\,$\pm$.10 & 0.44\,$\pm$.09 & 0.42\,$\pm$.09 & 0.39\,$\pm$.08 & 0.36\,$\pm$.08 & 0.34\,$\pm$.07 & 0.31\,$\pm$.06 \\
MV-Plurality       & \textbf{0.64}\,$\pm$.12 & \textbf{0.60}\,$\pm$.11 & 0.57\,$\pm$.11 & 0.54\,$\pm$.10 & 0.51\,$\pm$.09 & 0.48\,$\pm$.09 & 0.44\,$\pm$.08 & 0.41\,$\pm$.08 & 0.38\,$\pm$.07 & 0.35\,$\pm$.07 \\
MV-P-NoTB          & 0.63\,$\pm$.12 & 0.60\,$\pm$.11 & 0.56\,$\pm$.10 & 0.52\,$\pm$.10 & 0.48\,$\pm$.09 & 0.43\,$\pm$.08 & 0.39\,$\pm$.07 & 0.35\,$\pm$.06 & 0.30\,$\pm$.05 & 0.25\,$\pm$.04 \\
MV-Quorum          & 0.46\,$\pm$.12 & 0.44\,$\pm$.11 & 0.41\,$\pm$.11 & 0.39\,$\pm$.10 & 0.37\,$\pm$.10 & 0.34\,$\pm$.09 & 0.32\,$\pm$.08 & 0.30\,$\pm$.08 & 0.28\,$\pm$.07 & 0.25\,$\pm$.07 \\
MV-Strict          & 0.46\,$\pm$.12 & 0.44\,$\pm$.11 & 0.41\,$\pm$.11 & 0.39\,$\pm$.10 & 0.36\,$\pm$.10 & 0.34\,$\pm$.09 & 0.31\,$\pm$.08 & 0.29\,$\pm$.08 & 0.26\,$\pm$.07 & 0.24\,$\pm$.07 \\
IP                 & 0.63\,$\pm$.12 & 0.60\,$\pm$.11 & 0.57\,$\pm$.11 & 0.55\,$\pm$.10 & 0.52\,$\pm$.10 & 0.50\,$\pm$.10 & 0.48\,$\pm$.09 & 0.46\,$\pm$.09 & 0.44\,$\pm$.09 & 0.42\,$\pm$.08 \\
IP+TB              & 0.63\,$\pm$.12 & 0.60\,$\pm$.11 & \textbf{0.57}\,$\pm$.11 & \textbf{0.55}\,$\pm$.10 & \textbf{0.52}\,$\pm$.10 & \textbf{0.50}\,$\pm$.10 & \textbf{0.48}\,$\pm$.09 & \textbf{0.46}\,$\pm$.09 & \textbf{0.44}\,$\pm$.09 & \textbf{0.42}\,$\pm$.08 \\
HS                 & 0.63\,$\pm$.12 & 0.60\,$\pm$.11 & 0.57\,$\pm$.11 & 0.54\,$\pm$.10 & 0.50\,$\pm$.09 & 0.48\,$\pm$.09 & 0.45\,$\pm$.08 & 0.42\,$\pm$.08 & 0.39\,$\pm$.07 & 0.36\,$\pm$.07 \\
HS+TB              & 0.64\,$\pm$.12 & 0.60\,$\pm$.11 & 0.57\,$\pm$.11 & 0.55\,$\pm$.10 & 0.52\,$\pm$.10 & 0.49\,$\pm$.10 & 0.46\,$\pm$.09 & 0.43\,$\pm$.09 & 0.40\,$\pm$.08 & 0.37\,$\pm$.08 \\
\bottomrule
\end{tabular}%
}
\end{table*}

\noindent\textbf{Observations.} Four points worth noting on Table~\ref{tab:adv_agg}.

\noindent\emph{(i) Effect of TB on MV-Plurality.} On clean data MV-P and MV-P-NoTB are indistinguishable ($0.64$ vs.\ $0.63$). Under attack the gap widens monotonically:  at $p=0.9$ MV-P holds at $0.35$ F1 while MV-P-NoTB collapse to $0.25$. The $0.10$ F1 difference is entirely due to the confidence tie-break, which rescues plurality from the random ties induces by the coordinate flips.

\noindent\emph{(ii) Effect of TB on our methods.} IP and IP+TB are virtually identical at every $p$ (e.g., $0.422$ vs.\ $0.423$ at $p=0.9$): the abduction step already leaves almost no residual ambiguity to resolve. HS+TB beats raw HS by $\sim 0.015$ F1 at $p=0.9$, because the greedy search admits slightly more ambiguous (model, class) pairs that TB then disambiguates.

\noindent\emph{(iii) Where the crossover happens.} On clean data MV-Plurality is marginally ahead of IP+TB in the aggregate ($0.636$ vs.\ $0.632$). From $p\!\ge\!0.3$ IP+TB takes the lead, and the gap widens monotonically, reaching $+0.07$ F1 ($+22\%$ relative) at $p=0.9$.

\noindent\emph{(iv) Conservative MV variants and per-model baselines.} Macro AVG, MV-Quorum and MV-Strict stay below Best Model at every $p$ in the aggregate, and never come within $0.10$ F1 of either MV-Plurality or our methods. Their precision bias offers no robustness advantage under this attack.

\subsection{F. Running-Time Analysis}
\label{app:time}
\begin{figure}
    \centering
    \includegraphics[width=\linewidth]{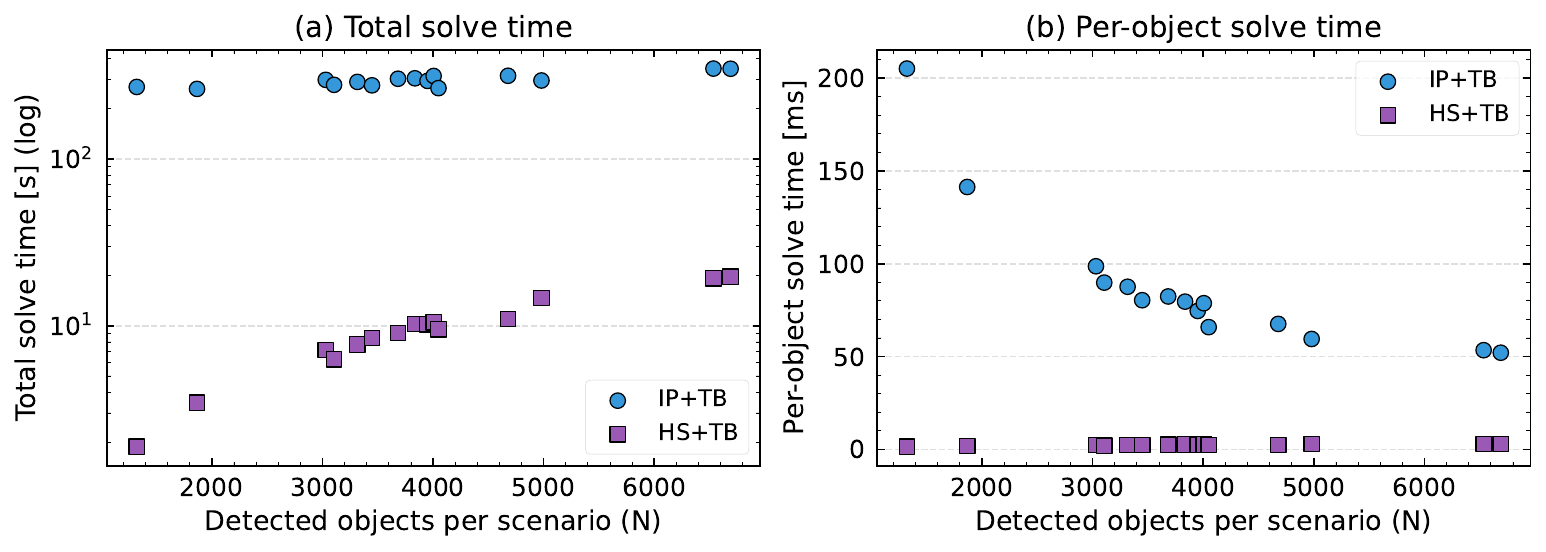}
    \caption{Empirical runtime of IP and HS as a function of the number of objects per images. \textit{(left)} total solve time and \textit{(right)} per-object solve time}
    \label{fig:time_analysis_app}
\end{figure}

Figure~\ref{fig:time_analysis_app} shows the empirical runtime of IP and HS as a function of the number of objects per images and the total solver runtime. On the left, we seee the total runtime, demonstrating the advantages of HS+TB, and the right the runtime per object per image, showing the convergence of IP+TB, consistent with the $O(N|\mathcal{F}||\mathcal{C}|)$ analysis.

\subsection{G. Combining LVP-EDR + DK-EDR}
\label{app:lvp_and_or_dk_edr}
To explore potential synergies between our EDR rules approaches, we evaluated combining the outputs of LVP-EDR and DK-EDR using Union ($LVP \cup DK$) and Intersection ($LVP \cap DK$) operations. However, results across the five clean test sets demonstrated that no combination outperforms the individual rules, with either LVP or DK alone achieving the highest F1 scores in all evaluated scenarios. For instance, in the AM\_1 dataset using the IP+TB configuration, the maximum F1 score reached is 0.342 for LVP and 0.344 for DK, whereas the Union reduces performance to 0.330 and the Intersection decreases it to 0.329 (Figure~\ref{fig:performance_lvp_dk_app}). This widespread performance degradation---with drops of up to -0.013 compared to LVP alone---indicates that the downstream abductive step is already operating at its maximum selection capacity; therefore, forcing additional filters ends up discarding valid predictions instead of improving the final result.

\begin{figure}
    \centering
    \includegraphics[width=\linewidth]{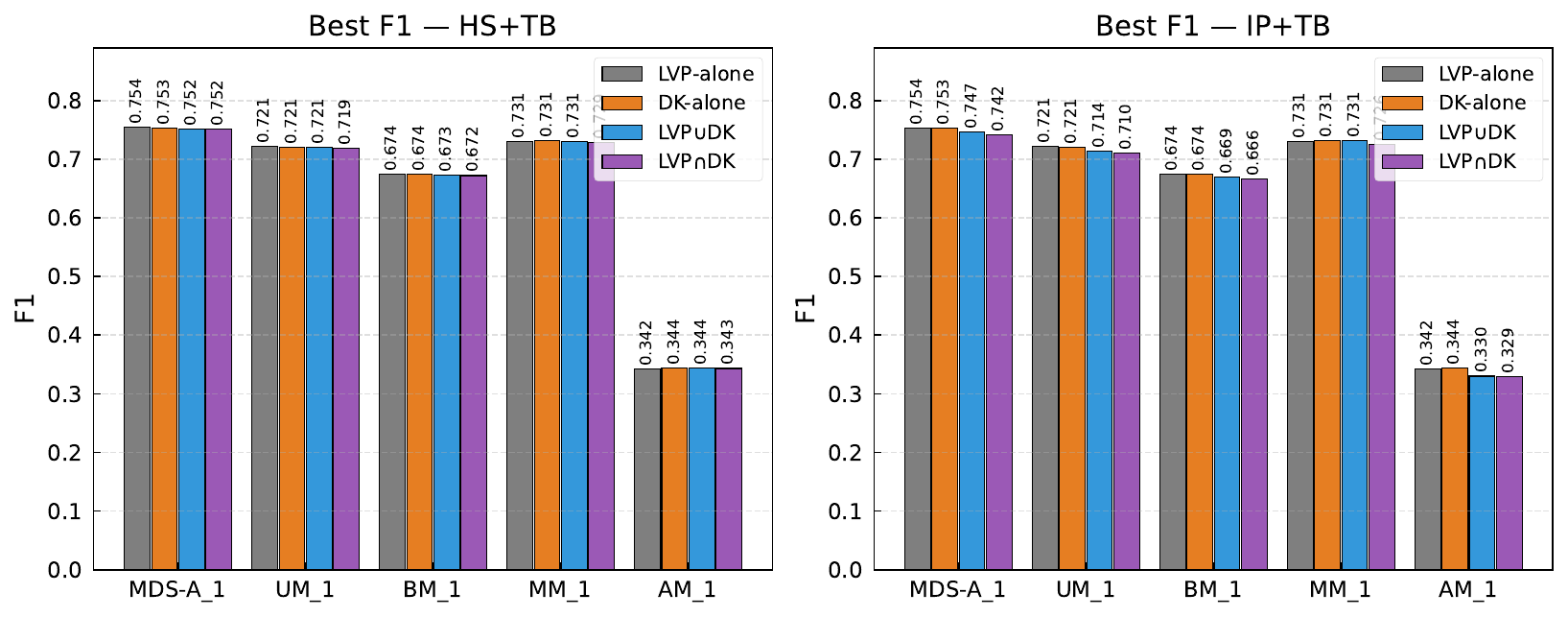}
    \caption{Comparison of maximum F1 scores across five clean test sets for individual detection methods (LVP-alone and DK-alone) versus their logical combinations (Union, $LVP \cup DK$, and Intersection, $LVP \cap DK$). Results are shown for both HS+TB and IP+TB abductive steps. In all evaluated scenarios, the combined filters fail to outperform the best standalone method, indicating a performance degradation during the downstream abduction.}
    \label{fig:heatmap_lvp_dk_app}
\end{figure}

\begin{figure}
    \centering
    \includegraphics[width=\linewidth]{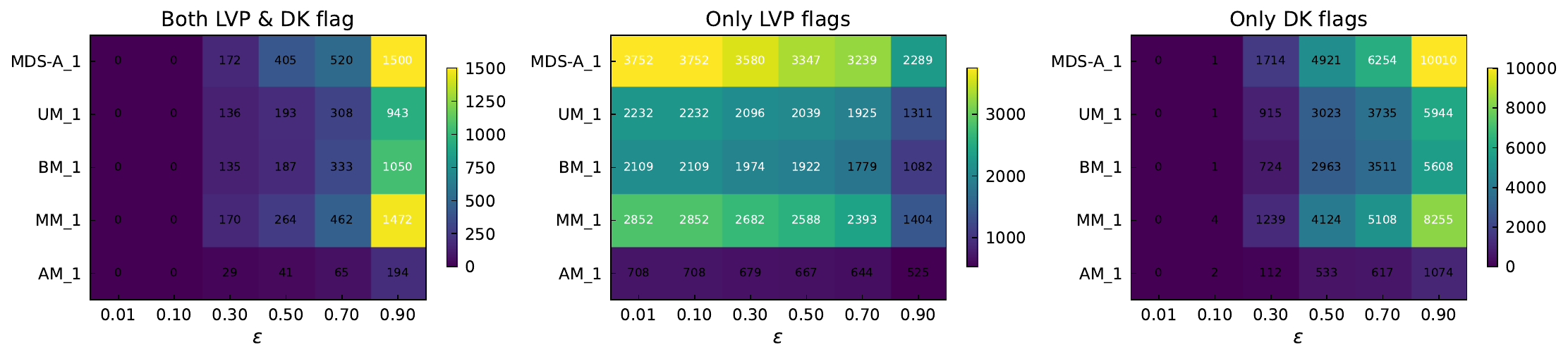}
    \caption{Detection overlap between LVP-EDR and DK-EDR across varying rule rigor levels ($\epsilon$) for the five evaluated test sets. The matrices display the number of errors flagged simultaneously by both methods, solely by LVP, and solely by DK. The identified error sets are nearly disjoint (e.g., yielding only a 4.7\% overlap in MDS-A\_1 at $\epsilon=0.5$), showing that while the methods are fundamentally complementary, combining them creates an overly aggressive filter that hinders the abductive step.}
    \label{fig:performance_lvp_dk_app}
\end{figure}

The overlap analysis reveals that this failure occurs because both methods identify almost completely disjoint sets of errors, operating on fundamentally different principles (LVP focuses on anomalies in the embedding space, while DK relies on geometry and masks). The confusion matrices show that, for the MDS-A\_1 scenario at $\epsilon=0.5$, only 405 detections are simultaneously flagged by both methods, compared to 3347 identified exclusively by LVP and 4921 exclusively by DK, representing a mere 4.7\% overlap (Figure~\ref{fig:heatmap_lvp_dk_app}). Although this confirms a strong theoretical complementarity between the approaches, stacking such divergent error signals under the current abductive combiner tightens the filter too aggressively. Given that straightforward logical combinations fail to exploit this complementarity without penalizing the F1 score, their evaluation on the remaining test suites and adversarial scenarios was discarded.

\subsection{H. External-Validity Experiment: Multimodal Test}
\label{app:external:drone}
To assess how the LVP-EDR framework transfers beyond the MDS-A dataset, we ran the full pipeline on VisDrone-DroneVehicle~\cite{sun2020drone}, a multimodal aerial dataset with paired RGB and infrared modalities. This setup subjects the framework to a fundamentally different regime than MDS-A: only two detectors (one per modality) instead of six weather-specific models. We first present the dataset and setup, then the results with clean and adversary data, and conclude with a discussion of when the abduction layer adds value.

\noindent\textbf{Dataset and Setup.}
\label{app:external:drone:setup}
VisDrone\nobreakdash-DroneVehicle~\cite{sun2020drone} provides $28{,}439$ aligned RGB\,/\,infrared aerial image pairs of urban and suburban scenes with five vehicle classes: \emph{car}, \emph{truck}, \emph{bus}, \emph{van}, and \emph{freight\_car}. Annotations are oriented bounding boxes; for our pipeline we convert them to axis-aligned bounding rectangles, strip the $100$-pixel annotation border built into the released files (training resolution becomes $640\times512$), and randomly sample $2000$ train + $200$ val + $500$ test pairs (seed $42$) to keep compute tractable while retaining full class coverage.

To make the ensemble evaluation meaningful, we build a \emph{mixed test set} of $500$ images composed of $250$ RGB scenes and $250$ IR scenes drawn from disjoint physical locations, so that each detector faces both in-domain and out-of-domain inputs at test time (see Fig.~\ref{fig:drone:gt} for a sample pair).

\begin{figure}[t]
  \centering
  \begin{tabular}{cc}
    \includegraphics[width=0.45\linewidth]{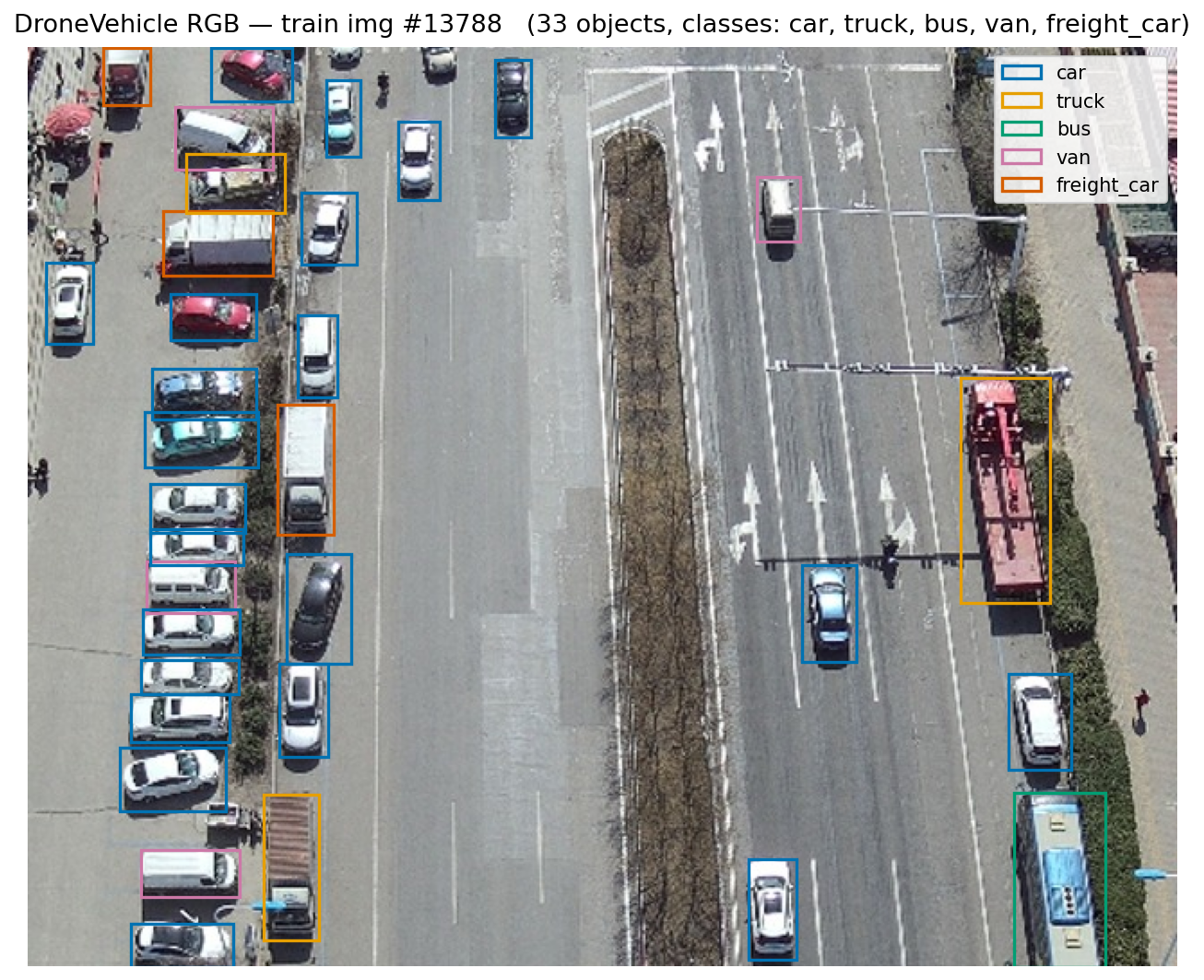} &
    \includegraphics[width=0.44\linewidth]{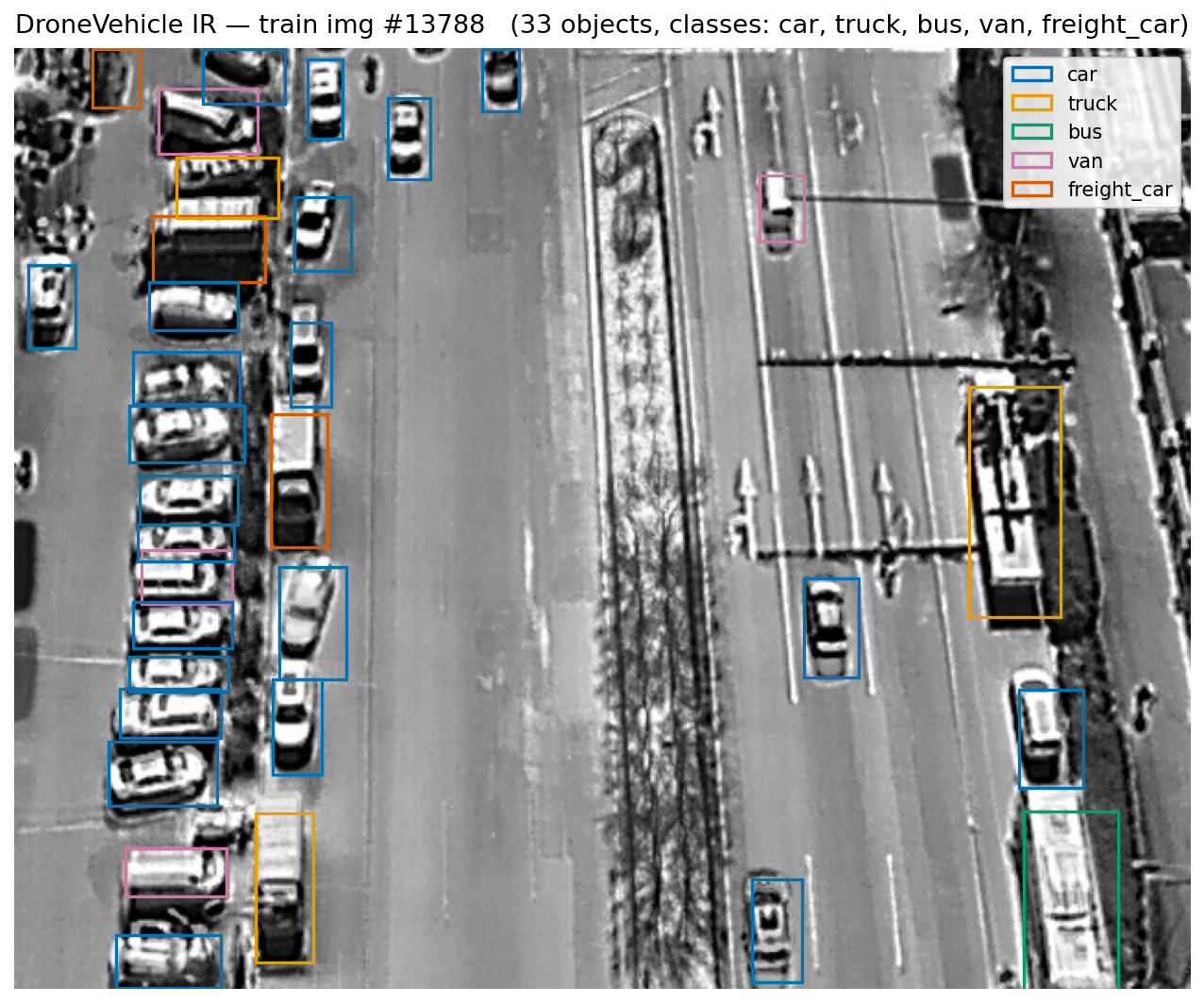} \\
  \end{tabular}
  \caption{DroneVehicle sample scene from the same physical location in
    RGB (left) and IR (right), with ground-truth bounding boxes
    color-coded by class.}
  \label{fig:drone:gt}
\end{figure}

\noindent\textbf{Detection Performance}
\label{app:external:drone:val}
Both detectors reach comparable $\text{AP}_{50}$ around $32$ on their respective validation splits (Table~\ref{tab:drone:val_ap}). IR is consistently equal or better than RGB on the well-represented classes, with the largest gap on \emph{car} where the thermal signature of engines gives IR a clear advantage. Van and freight\_car collapse to $\text{AP}=0$ in both modalities: with $\sim 900$ training instances and strong visual similarity to trucks in an aerial view, the models rarely commit to these ambiguous shapes. Downstream the pipeline therefore operates on three effective classes (car, truck, bus).

\begin{table}[t]
  \centering
  \small
  \setlength{\tabcolsep}{4pt}
  \begin{tabular}{lccccccc}
    \toprule
    Detector & AP & AP$_{50}$ & car & truck & bus & van & freight \\
    \midrule
    RGB & 18.8 & 32.0 & 48.3 & 9.2 & 36.4 & 0.0 & 0.0 \\
    IR  & 20.5 & 31.4 & 57.6 & 7.0 & 37.7 & 0.0 & 0.0 \\
    \bottomrule
  \end{tabular}
  \caption{DroneVehicle validation AP by class.}
  \label{tab:drone:val_ap}
\end{table}

\noindent\textbf{Baselines and Abduction on Clean Data}
\label{app:external:drone:clean}
Table~\ref{tab:drone:clean} reports the FULL-metric performance on the mixed test set (FP counts include false-positive detections without a matching GT). The best single detector is IR at $F_1 = 0.680$; the ensemble baselines cluster around $F_1 = 0.67$, with MV-Quorum penalized for enforcing agreement across the mixed in-domain/out-of-domain regime. Recall crashes without a proportional precision gain, unlike in MDS-A where MV-Quorum benefits from six independent votes.

Our IP+TB and HS+TB tie MV-Plurality at $F_1 = 0.672$ because the inter-detector inconsistency on the clean test set is essentially zero ($0.0002$). The abductive layer has no conflicts to resolve on clean data: RGB and IR were trained on paired scenes and their predictions on any given object almost always agree at the class level, so the IP/HS optimizers degenerate to the majority-vote solution. This is the worst-case regime for our framework in clean conditions, and yet, as the adversarial results show below, it becomes the best-case regime under attack.

\begin{table}[t]
  \centering
  \small
  \begin{tabular}{lccc}
    \toprule
    Method & Precision & Recall & $F_1$ \\
    \midrule
    RGB (single)         & 0.841 & 0.514 & 0.638 \\
    IR  (single)         & 0.871 & 0.558 & 0.680 \\
    Macro AVG            & 0.856 & 0.536 & 0.659 \\
    Best Model (= IR)    & 0.871 & 0.558 & \textbf{0.680} \\
    MV-Plurality         & 0.785 & 0.588 & \textbf{0.672} \\
    MV-Plurality-NoTB    & 0.785 & 0.588 & \textbf{0.672} \\
    MV-Quorum            & 0.962 & 0.484 & 0.644 \\
    MV-Strict            & 0.962 & 0.484 & 0.644 \\
    IP+TB (ours)         & 0.792 & 0.584 & \textbf{0.672} \\
    HS+TB (ours)         & 0.792 & 0.584 & \textbf{0.672} \\
    \bottomrule
  \end{tabular}
  \caption{DroneVehicle mixed-test performance on clean data
    (FULL metric). Inter-detector inconsistency is $0.0002$; IP+TB and
    HS+TB degenerate to MV-Plurality because there are no conflicts to
    resolve.}
  \label{tab:drone:clean}
\end{table}

\noindent\textbf{Adversarial Coordinated Label-Flip}
\label{app:external:drone:adv}
We apply the same coordinated label-flip protocol used in the main paper. For each ground-truth object, with probability $p$, we pick one of the two detectors ($k=1$: partial compromise) and replace its predicted class with a random wrong class. We sweep $p \in \{0, 0.1, \ldots, 0.9\}$ with three adversary seeds and reuse the LVP classifiers and EDR rules learned on clean data (no retraining per attack).

\begin{figure}[t]
  \centering
  \begin{tabular}{cc}
    \includegraphics[width=0.47\linewidth]{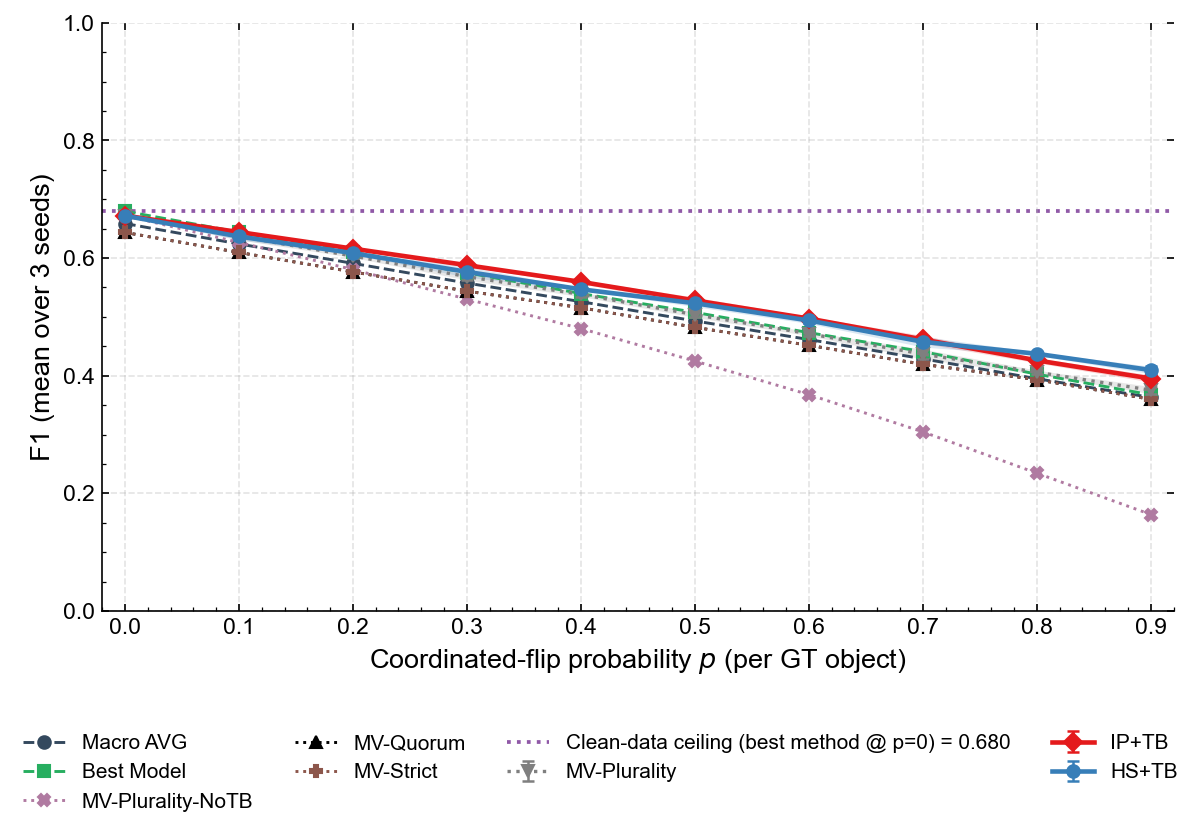} &
    \includegraphics[width=0.46\linewidth]{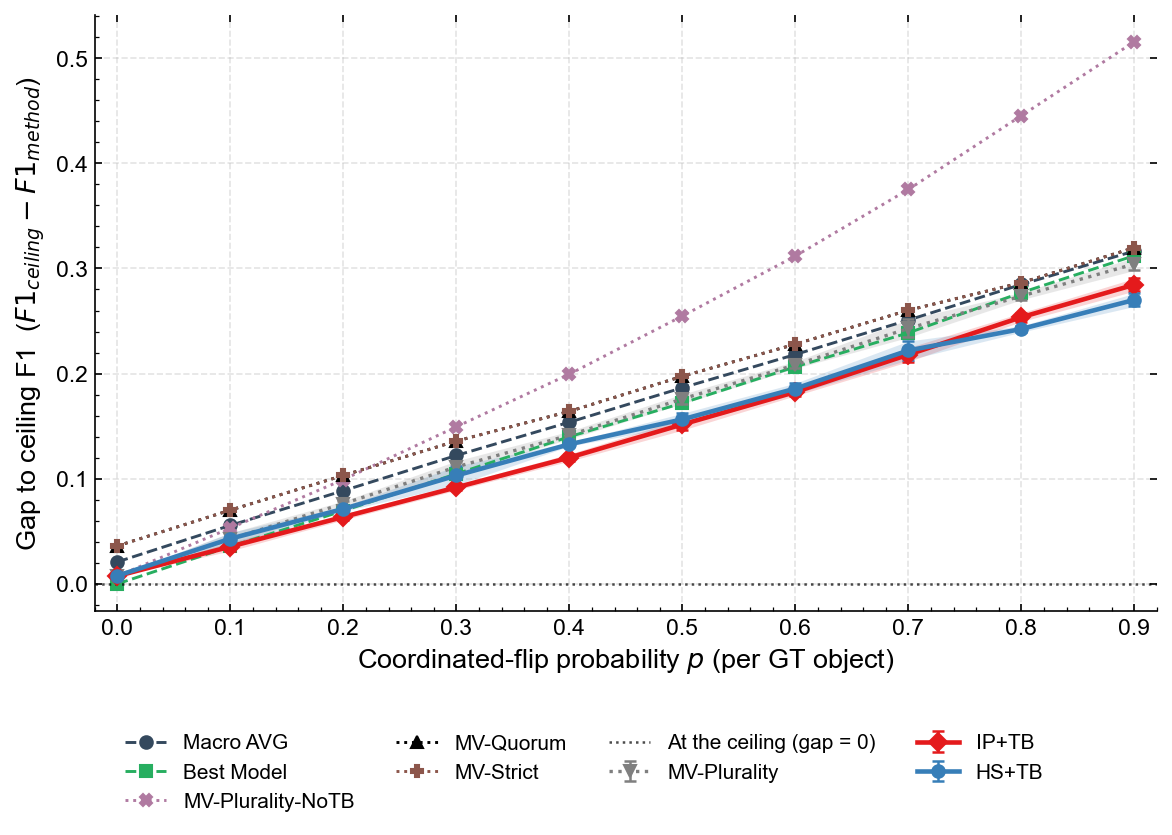} \\
  \end{tabular}
  \caption{(Left:) Adversarial coordinated label-flip ($k=1$):
    $F_1$ vs.\ attack probability $p$, averaged over three adversary
    seeds. Error bars are $\pm 1$~std on the three methods
    (MV-Plurality, IP+TB, HS+TB). IP+TB (red) and HS+TB (blue) are the
    top two lines for all $p \ge 0.1$; MV-Plurality-NoTB collapses. (Right:) Gap to the clean-data ceiling
    ($F_1^{\text{ceiling}}=0.680$, achieved by the IR detector alone at
    $p=0$) as a function of attack probability. IP+TB and HS+TB stay
    closest to the ceiling across the sweep; MV-Plurality-NoTB grows
    fastest.}
  \label{fig:drone:adv}
\end{figure}

Both of our unconstrained methods \emph{stay above every baseline} for all $p \ge 0.1$ (see left in Fig.~\ref{fig:drone:adv}). At $p=0.9$, IP+TB reaches $F_1=0.395$ and HS+TB $F_1=0.410$, versus MV-Plurality $0.376$, MV-Quorum $0.360$, and Best Model $0.367$; MV-Plurality-NoTB collapses to $F_1=0.16$ because the adversary-induced ties trigger its abstention policy. The gap to the clean-data ceiling widens for every method as $p$ grows, but stays smallest for IP+TB and HS+TB across the whole sweep (see right in Fig.~\ref{fig:drone:adv}). The coordinated flip introduces the inter-detector inconsistency that the abduction step was designed to resolve.

\noindent\textbf{Discussion}
\label{app:external:drone:discussion}
DroneVehicle exercises the LVP-EDR framework in a regime that is strictly harder than MDS-A for the abductive layer to add value on clean data: only two detectors, and both trained on paired scenes with high behavioral correlation. Inter-detector inconsistency is essentially zero and no ensemble method, ours included, meaningfully beats the best single detector (IR at $F_1=0.680$). The clean-data story is a tie between our methods and MV-Plurality, and a mild loss against Best Model.

The adversarial results reverse this picture. The coordinated label-flip attack \emph{creates} the very inconsistency the abductive layer requires, and the LVP filter, learned from clean data alone, flags the flipped detections. As a result, both IP+TB and HS+TB stay every baseline across the full attack sweep, with a margin that widens with attack strength (up to $\sim 5$~$F_1$ points over MV-Quorum at $p=0.9$).

We view this as a complementary corroboration of the paper's central claim: the LVP metacognitive layer adds most value precisely when the base ensemble faces disagreements it cannot resolve on its own. On MDS-A those disagreements arise naturally from six specialized weather detectors and yield strong clean-data wins; on DroneVehicle they arise only under adversarial pressure and yield strong adversarial wins.
\end{document}